\def\paperlanguage{} %% English
\pdfoutput=1 % for arXiv

\documentclass[letterpaper, 10 pt, conference]{ieeeconf}  % Comment this line out if you need a4paper

\usepackage{bm}
\usepackage{cite}
\usepackage{flushend}
\usepackage{here}
\usepackage{multirow}
\input{preamble}

\IEEEoverridecommandlockouts                              % This command is only needed if
\title{
  {
    \switchlanguage%
    {%
    A Universal Wire Testing Machine \\for Enhancing the Performance of Wire-Driven Robots
    }%
    {%
    A Universal Wire Testing Machine \\for Enhancing the Performance of Wire-Driven Robots
    }%
  }
}

\author{Temma Suzuki$^{1}$, Kento Kawaharazuka$^{1}$, and Kei Okada$^{1}$
  \thanks{
    $^{1}$ The authors are with the Department of Mechano-Informatics, Graduate School of Information Science and Technology, The University of Tokyo, 7-3-1 Hongo, Bunkyo-ku, Tokyo, 113-8656, Japan.
    {\text\small [t-suzuki, kawaharazuka, k-okada]@jsk.imi.i.u-tokyo.ac.jp}
  }
}
\begin{document}

\maketitle

%%%%%%%%%%%%%%%%%%%%%%%%%%%%%%%%%%%%%%%%%%%%%%%%%%%%%%%%%%%%%%%%%%%%%%%%%%%%%%%%
\begin{abstract}
  \switchlanguage%
  {%
    Compared with gears and linkages, wires constitute a lightweight, low-friction transmission mechanism. 
    However, because wires are flexible materials, they tend to introduce large modeling errors, and their adoption in industrial and research robots remains limited. 
    In this study, we built a Universal Wire Testing Machine that enables measurement and adjustment of wire characteristics to improve the performance of wire-driven mechanisms. 
    Using this testing machine, we carried out removal of initial wire stretch, measurement of tension transmission efficiency for eight different diameters of passive pulleys, and measurement of the dynamic behavior of variable-length wires. 
    Finally, we applied the data obtained from this testing machine to the force control of an actual wire-driven robot, reducing the end-effector force error.
  }%
  {%
    ワイヤは歯車やリンクと比較して, 軽量で低摩擦な動力伝達機構である.
    一方でワイヤは柔軟な素材であるためモデル化誤差が大きく, 産業・研究用ロボットでの採用は少ない.
    本研究ではワイヤ駆動ロボットの性能向上に向けて, ワイヤ特性の測定と調節が可能なUniversal Wire Testing Machineを製作した.
    本試験機を用いて, ワイヤの初期伸びの除去, 8種類の直径の受動プーリにおける張力伝達効率の測定, 可変長ワイヤの動特性の測定を行った.
    さらに本試験機で得たデータを実際のワイヤ駆動ロボットの力制御に応用し, 手先力の誤差を低減した.
  }%
\end{abstract}

\section{Introduction}\label{sec:introduction}
\switchlanguage%
{%
Compared with gears and linkages, wires offer a low-friction, low-inertia transmission mechanism. 
In robots requiring safe interaction with their environments, such as collaborative robot arms, wire-driven mechanisms have often been employed. 
The collaborative robot arm WAM Arm realized high backdrivability and zero backlash by transmitting motor power via wires \cite{wamarm2024}. 
In the LIMS series of robot arms, the actuators that drive the wrist are consolidated in the upper arm through a wire-driven mechanism, achieving low forearm inertia \cite{kim2017anthropomorphic, song2018lims2}.

The coupled wire-driven mechanism, a type of wire-driven transmission mechanism, transmits power using a number of wires that is redundant relative to the number of joints \cite{hirose1991tendon, yokoi1991design, suzuki2024saqiel, suzuki2022ramiel, inoue2025overcoming, yoshimura2023design}. 
Using the coupled wire-driven mechanism allows the motors to be consolidated on the root link, minimizing the inertia of the moving parts. 
The coupled wire-driven robot arm SAQIEL achieved link lengths and output forces equivalent to those of a human arm at a mass of less than half that of a human arm (\SI{1.5}{\kilogram}) \cite{suzuki2024saqiel}. 
Another advantage of the coupled wire-driven mechanism is its high design freedom of the achievable joint torque space \cite{kawaharazuka2023design}. 
In SAQIEL, the shoulder-roll and shoulder-pitch joints are driven by power equivalent to that of ten motors, thereby achieving both weight reduction and payload capacity \cite{suzuki2024saqiel}.

\begin{figure}[t]
  \centering
  \includegraphics[width=1.0\columnwidth]{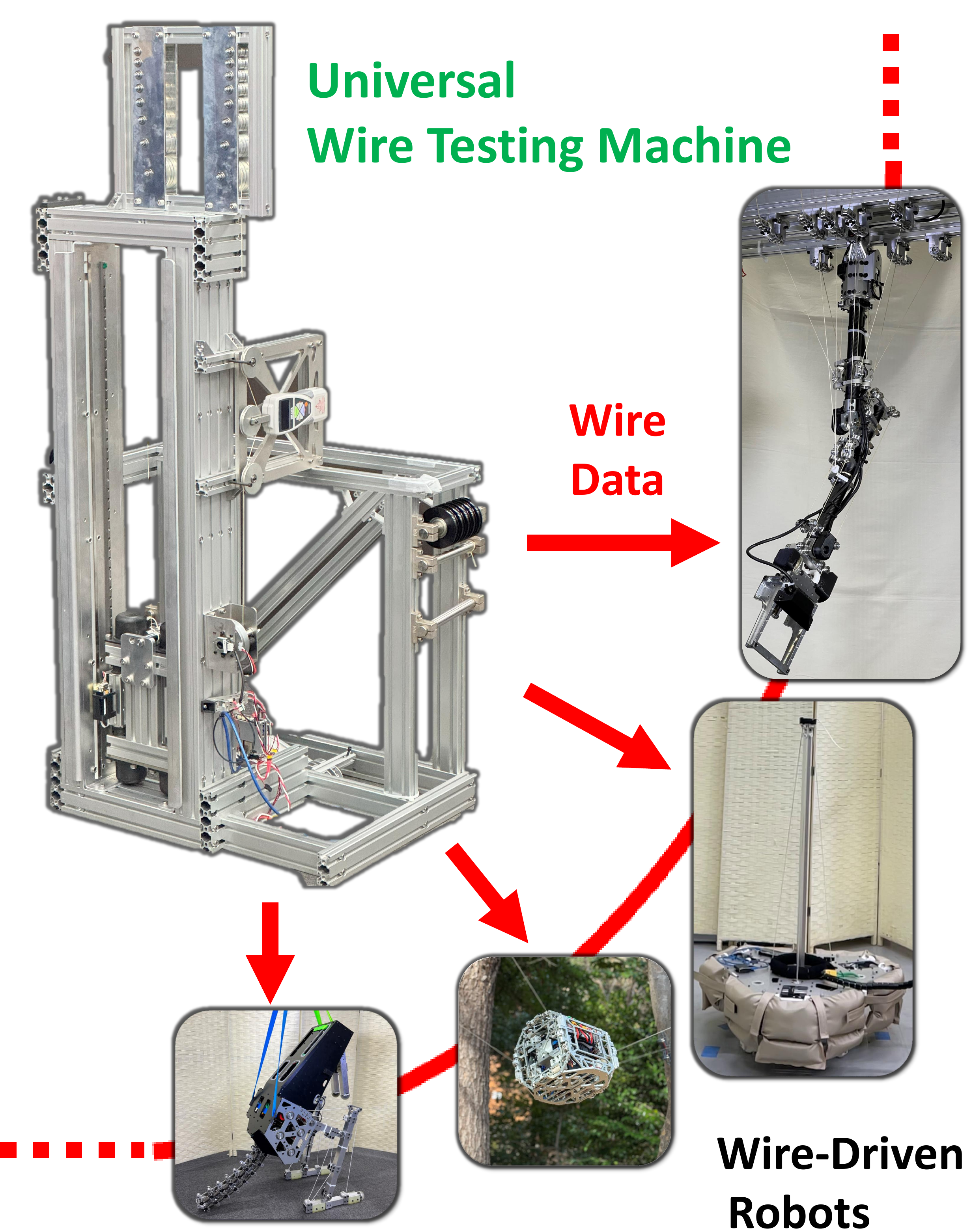}
  \vspace{-2ex}
  \caption{Overview of universal wire testing machine.}
  \label{figure:wiretester}
  \vspace{-1.5ex}
\end{figure}

Despite various advantages of wire-driven mechanisms, including lightweight construction, low friction, and high design freedom, most industrial and research robots employ transmission mechanisms based on gears or linkages. 
A major reason why wire-driven mechanisms have not become widespread is that three issues tend to increase modeling error compared with other transmission mechanisms. 
First, wires are flexible mechanical elements compared with gears or linkages and thus undergo initial wire stretch due to plastic deformation shortly after first use. 
Second, when a wire passes through a passive pulley, loss of tension occurs, reducing tension transmission efficiency \cite{miyasaka2015measurement, mate2024measurement}. 
Third, wires exhibit viscoelasticity and therefore display hysteresis and oscillatory behavior under dynamic loading \cite{takata2018modeling}. 
These wire properties introduce nonlinearities in wire-driven mechanisms that hinder control.

Therefore, this study proposes a Universal Wire Testing Machine (\figref{figure:wiretester}) for improving the performance of wire-driven robots by enabling measurement and adjustment of wire characteristics. 
The testing machine has the following three functions: 1) removal of initial wire stretch, 2) measurement of tension transmission efficiency in a passive pulley, and 3) measurement of tension behavior under dynamic loading. 
We conduct experiments to validate these three functions. 
Furthermore, using the obtained data on tension transmission efficiency, we improve the force-control accuracy of an actual wire-driven robot. 
By adjusting and measuring various wire properties with these three functions, this work contributes to enhancing the performance of wire-driven robots.
}%
{
ワイヤは歯車やリンクと比較して, 低摩擦, 低慣性な動力伝達機構である.
協働ロボットアームのような環境との安全なインタラクションが求められるロボットではしばしワイヤ駆動機構が採用されてきた.
協働ロボットアームWAM Armは, ワイヤでモータの動力を伝達することで, 高いバックドライバビリティとゼロバックラッシュを実現した\cite{wamarm2024}.
ロボットアームLIMSシリーズではワイヤ駆動によって手首を駆動するアクチュエータを上腕に集約し, 前腕の低慣性化を達成した\cite{kim2017anthropomorphic, song2018lims2}.

ワイヤ駆動の一種であるワイヤ干渉駆動機構は, 関節数に対して冗長な本数のワイヤを用いて動力を伝達する機構である\cite{hirose1991tendon, yokoi1991design, suzuki2024saqiel, suzuki2022ramiel, inoue2025overcoming, yoshimura2023design}.
ワイヤ干渉駆動機構を用いることでモータをルートリンクに集約し, 動作部の慣性を最小化することができる.
ワイヤ干渉駆動ロボットアームSAQIELは人間の腕と同等のリンク長, 発揮力を, 人間の腕の半分以下の質量(\SI{1.5}{\kilogram})で達成した\cite{suzuki2024saqiel}.
ワイヤ干渉駆動機構の利点としては発揮可能関節トルク空間の設計自由度が高いことも挙げられる\cite{kawaharazuka2023design}.
ロボットアームSAQIELではshoulder-roll jointと shoulder-pitch jointの駆動にモータ10個分の動力を用いており, 軽量化とペイロードの確保の両立を実現している\cite{suzuki2024saqiel}.
% 多くのロボットでは関節とモータが一対一対応しており, 各関節の発揮可能トルクは各モータの発揮可能トルクに比例している.
% 一方でワイヤ干渉駆動では一つの関節の駆動に複数のモータを用いるため, 関節の負荷に応じてモータ出力の配分を調整することが可能となる.
% そして各ワイヤが各関節で発生させるトルクは, ワイヤの経由点の位置変更や, 関節に同軸なプーリの直径を変更することで簡単に調節できる.

ワイヤ駆動には軽量, 低摩擦, 高い設計自由度といった様々な利点があるにもかかわらず, 産業・研究用途のロボットの殆どは歯車やリンクによる動力伝達を採用している.
ワイヤ駆動機構が普及しない大きな要因として, 以下の3つの問題によってモデル化誤差が他の駆動機構と比べて大きくなりやすいことが考えられる.
1)ワイヤは歯車やリンクと比べて柔軟な機械要素であるため, 使用開始直後の塑性変形が起こりやすい.
2)ワイヤが受動プーリを経由する際に, 張力の損失が発生する\cite{miyasaka2015measurement, mate2024measurement}.
3)ワイヤには粘弾性が存在するため, 動的負荷を与えるとヒステリシス挙動や振動的な挙動を示す\cite{takata2018modeling}.
このようなワイヤの性質によって, ワイヤ駆動機構には制御を妨げる非線形性が存在する. 

そこで本研究ではワイヤ駆動ロボットの性能向上に向けて, ワイヤ特性の測定と調節が可能なUniversal Wire Testing Machine(\figref{figure:wiretester})を提案する.
本試験機には次の3つの機能がある, 1)ワイヤの初期伸びの除去, 2)受動プーリにおける張力伝達効率の測定, 3)動的負荷を与えた場合の張力挙動の測定.
そして実際にこれら3つの機能のテストを行う.
さらに取得した張力伝達効率のデータを用いて, 実際のワイヤ駆動ロボットの力制御精度を向上させる.
本試験機の3つの機能を用いて多様なワイヤの特性を調節・測定し, ワイヤ駆動ロボットの性能向上に貢献する.

\begin{figure}[t]
  \centering
  \includegraphics[width=1.0\columnwidth]{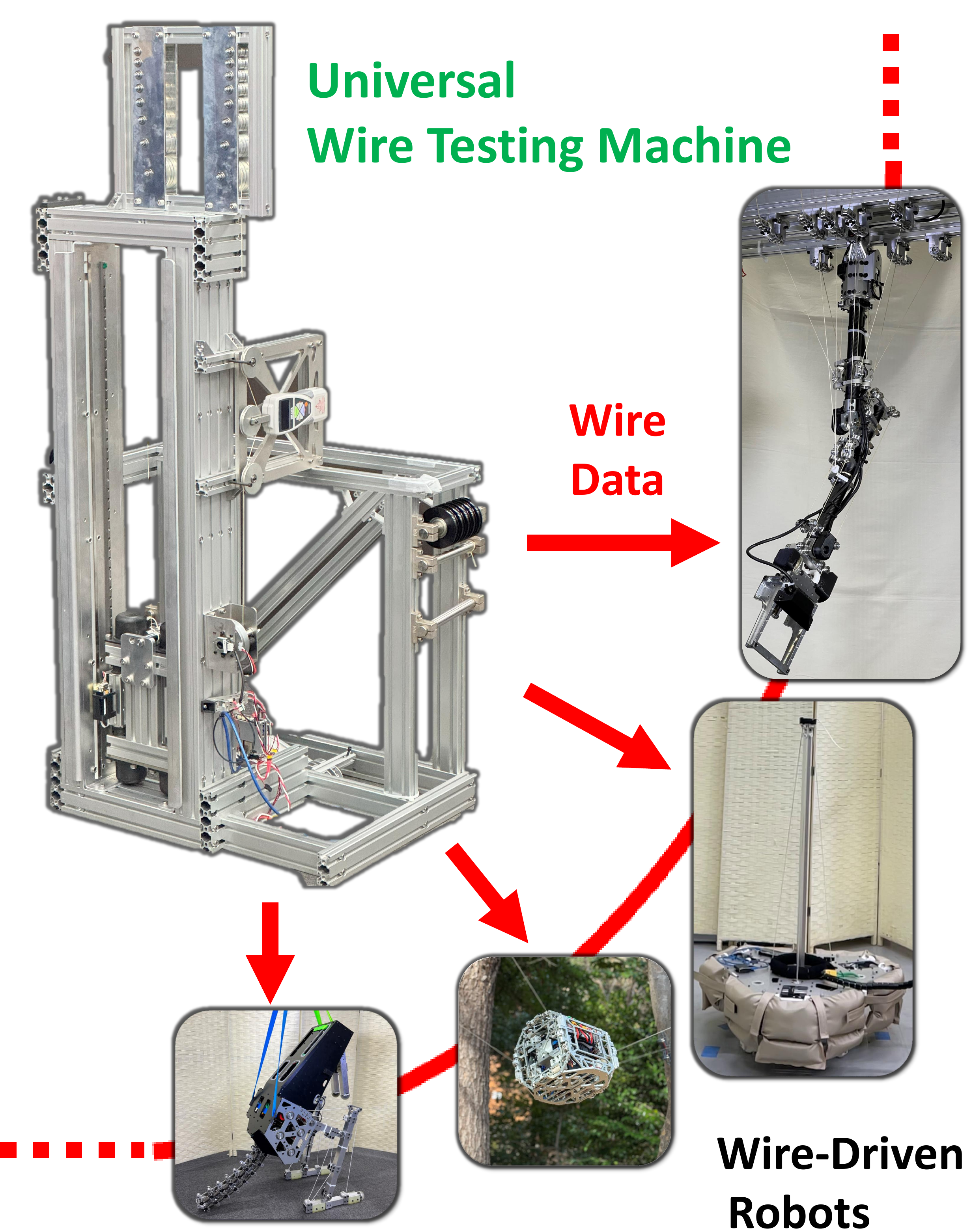}
  \vspace{-2ex}
  \caption{Overview of universal wire testing machine.}
  \label{figure:wiretester}
  \vspace{-1.5ex}
\end{figure}
}%

\section{Characteristics of the Wire-Driven Mechanism} \label{sec:characteristics}
\switchlanguage%
{%
Characteristics of wires that influence the control of wire-driven robots include plastic elongation, tension loss in passive pulleys, and viscoelasticity. 
This chapter provides a detailed investigation of these wire characteristics and establishes the required specifications for the Universal Wire Testing Machine.
}%
{%
ワイヤ駆動ロボットの制御に影響を及ぼすワイヤの特性として, plastic elognation, 受動プーリにおける張力損失, 粘弾性が挙げられる.
本章ではこれらのワイヤの特性について詳細な調査を行い, ワイヤ試験機の要求仕様を決定する.
}

\subsection{Plastic Elongation} \label{subsec:plastic-elongation}
\switchlanguage%
{%
Applying tension repeatedly to a new wire is known to cause plastic elongation. Therefore, to improve the reproducibility of the strain-tension relationship in wire-driven mechanisms, initial wire stretch must be removed by preloading the wire several times. 
In particular, synthetic fiber ropes, which have been widely adopted in recent wire-driven robots \cite{suzuki2024saqiel, endo2019superdragon}, are known to creep more readily than steel ropes and to exhibit stronger hysteresis in the strain-tension relationship. 
Previous research demonstrated that, under a static load of 75 N for one hour, steel ropes elongated by less than \SI{1}{\percent}, whereas Dyneema synthetic fiber ropes exhibited approximately \SI{15}{\percent} elongation \cite{gueners2021cable}. 
To enhance the reproducibility of operations in robots that transmit power via synthetic fiber or steel ropes, the Universal Wire Testing Machine is equipped with a function for removal of initial wire stretch.
}%
{%
新品のワイヤに何回か張力を印加すると, ワイヤが塑性変形し伸びることが知られている.
そのためワイヤ駆動ロボットで歪-張力の関係の再現性を高めるためには, 事前に何回か張力を印加し初期伸びを取り除く必要がある.
特に近年ワイヤ駆動ロボットで多く採用されいてる合成繊維ロープ\cite{suzuki2024saqiel, kawaharazuka2019musashi, endo2019superdragon}はスチールロープよりもクリープしやすく, また歪-張力関係におけるヒステリシス性も強いことが知られている.
先行研究ではスチールロープと合成繊維ロープdyneemaに1時間75Nの静荷重をかけたところ, スチールは\SI{1}{\percent}以下の伸びであった一方, dyneemaでは\SI{15}{\percent}ほどの延伸が発生したことが確認されている\cite{gueners2021cable}. 
そこで本研究では合成繊維ロープやスチールロープによって動力を伝達するロボットの動作の再現性を高めるために, ワイヤ試験機にワイヤの初期伸びを除去する機能を搭載する
}%

\subsection{Tension Loss in Wire Pulleys} \label{subsec:tension-loss}
\switchlanguage%
{%
It is known that tension decreases when a wire passes through a passive pulley \cite{miyasaka2015measurement, choi2017tension, mate2024measurement}. 
Measuring and modeling tension loss in passive pulleys is essential for improving the performance of wire-driven mechanisms. 
Miyasaka et al. investigated the relationship among cable velocity, tension, pulley type and number, wrap angle, and tension transmission efficiency for the combination of stainless steel wire and passive pulleys\cite{miyasaka2015measurement}. 
They showed that tension loss in pulleys is independent of wire velocity and depends on tension, wrap angle, and number of sheaves, and can be modeled as Coulomb friction \cite{miyasaka2015measurement}. 
Furthermore, Sung-Hyun et al. pointed out that friction not captured by the Coulomb friction model arises in passive pulleys when the wire reverses direction \cite{choi2017tension}. 
To represent the nonlinear friction occurring during direction changes, they adopted Dahl's friction model \cite{dahl1968friction} and demonstrated that it predicts tension more accurately than the Coulomb friction model in dynamic tasks \cite{choi2017tension}.

Here, the causes of tension loss when a wire passes through a passive pulley are presented. 
Mate et al. revealed that the majority of tension loss during wire passage through a passive pulley originates from friction between the material constituents inside the steel cable \cite{mate2024measurement}. 
They then modeled this tension loss due to internal cable friction as hysteric energy lost resulting from the cable's cyclic bending \cite{mate2024measurement}.

These previous studies have provided accurate measurements of tension loss for specific wire-pulley combinations. 
In practical wire-driven mechanisms, a wide variety of wire types and pulley diameters are employed to satisfy requirements such as size, range of motion, and stiffness. 
Consequently, promoting broader adoption of wire-driven mechanisms requires measurement of tension loss across a more diverse set of wire and pulley diameter combinations. 
The Universal Wire Testing Machine fabricated in this work measures tension transmission efficiency for multiple combinations of wire types and pulley diameters.
}%
{%
% [ワイヤが受動プーリを通過すると, 張力が減少する]
ワイヤが受動プーリを通過すると, 張力が減少することが知られている\cite{miyasaka2015measurement, choi2017tension, mate2024measurement}.
受動プーリにおける張力損失を計測しモデル化することは, ワイヤ駆動ロボットの性能向上に必要不可欠である.
miyasakaらは手術ロボットRAVEN IIで使用される, ステンレス鋼ワイヤと受動プーリの組み合わせについて, ケーブル速度, 張力, プーリの種類と数, プーリへのケーブルの巻き付け角度と張力伝達効率の関係を調べた\cite{miyasaka2015measurement}.
miyasakaらはプーリにおける張力損失が, ワイヤ速度とは関係がなく, 張力, 巻き付け角度, 滑車数の関数となっており, クーロン摩擦としてモデル化できることを示した\cite{miyasaka2015measurement}.
さらにSung-Hyunらはワイヤの動作方向を切り替える際に, クーロン摩擦モデルでは表現できない摩擦が受動プーリで発生することを指摘した\cite{choi2017tension}.
Sung-Hyunらは方向切り替え時に発生する非線形な摩擦を表現するためにDahlの摩擦モデル\cite{dahl1968friction}を採用し, 動的なタスクにおいてクーロン摩擦モデルより精度良く張力を予測できることを示した\cite{choi2017tension}.

ここで受動プーリを通過する際の張力損失が発生する原因について紹介する.
Mateらは鋼ワイヤが受動プーリを通過する際の張力損失の大部分がワイヤ内部の素材同士の摩擦によって発生していることを明らかにした\cite{mate2024measurement}.
そしてこのケーブルの内部摩擦による張力損失を, ケーブルの周期的な曲げによるhysteric energy lostとしてモデル化した\cite{mate2024measurement}.

これらの先行研究によって特定のワイヤ, プーリの組み合わせにおける張力損失が精度良く計測されてきた.
一方で実際のワイヤ駆動ロボットではサイズ, 可動域, 剛性などの要求に合わせて多様な種類のワイヤと多様な直径のプーリが使用される.
そのためワイヤ駆動ロボットを普及させるためには, より多様なワイヤと多様なプーリ直径の組み合わせにおける張力損失を計測する必要がある.
そこで本研究で製作するワイヤ試験機では, 複数種類のワイヤ, プーリの組み合わせにおける張力伝達効率を測定する.
}%

\subsection{Dynamic Characteristics of Wires} \label{subsec:dynamic-chracteristics}
\switchlanguage%
{%
The wire-driven mechanism is a low-stiffness transmission mechanism compared with metallic gears and linkages. 
To improve the accuracy of wire-driven robots and extend their usable frequency bandwidth, efforts have been made to measure and model the dynamic behavior of wires \cite{gueners2021cable, takata2018modeling}. 
Güners et al. measured the viscoelasticity of synthetic fiber and steel ropes using a tensile testing machine to assess the influence of wire behavior on cable-driven parallel robots (CDPRs) \cite{gueners2021cable}. 
They then represented the wires with a Kelvin-Voigt model comprising parallel elastic and viscous elements and simulated CDPR performance.

Takata et al. constructed a long-distance cable-pulley system in which a servomotor and a fly wheel are connected by a \SI{15.2}{\metre} wire to investigate the frequency response of a wire-driven mechanism in a decommissioning robot arm \cite{takata2018modeling}. 
Using this apparatus, they measured the frequency response of synthetic fiber ropes and stainless steel ropes, and represented wire behavior with a four-element model composed of two elastic elements and one viscous element \cite{takata2018modeling}.

These previous studies have provided detailed investigations into the dynamic behavior of wires under the assumption of a constant total length. 
However, in an actual wire-driven robot \cite{suzuki2024saqiel}, the total wire length may change by more than \SI{50}{\percent} depending on the robot's posture. 
To model the behavior of wire-driven robots accurately, it is necessary to measure the tension and strain of the wire when its total length undergoes significant changes. 
Therefore, the wire testing machine developed in this study enables the measurement of wire tension under dynamic variations in total length.
}%
{%
ワイヤ駆動機構は金属製の歯車やリンクといった機構と比べて低剛性な動力伝達機構である.
ワイヤ駆動ロボットの精度向上や動作可能周波数帯域の拡張のために, ワイヤの動的挙動を計測しモデル化する試みがなされてきた\cite{gueners2021cable, takata2018modeling}.
Genersらはワイヤの挙動がcable-driven parallel robots(CDPR)に与える影響を検討するために, 引張試験機で合成繊維ロープとスチールロープの粘弾性を計測した\cite{gueners2021cable}.
そして並列な弾性要素と粘性要素を有するKelvin-Voigtモデルでワイヤを表現し, CDPRの挙動をシミュレーションした.

Takataらは廃炉作業用ロボットアームにおけるワイヤ駆動機構の周波数応答を調査するために, サーボモータとfly wheelが\SI{15.2}{\metre}のワイヤで接続されたLong-distance cable-pulley systemを構築した\cite{takata2018modeling}.
そしてこの装置を用いて合成繊維ロープとステンレスロープの周波数応答を計測し, ワイヤの挙動を2つの弾性要素と粘性要素からなる4要素モデルで表現した\cite{takata2018modeling}.

これらの先行研究によってワイヤの全長が一定な場合の動的挙動についての詳細な調査が行われてきた.
一方で実際のワイヤ駆動ロボット\cite{suzuki2024saqiel}ではロボットの姿勢に応じてワイヤの全長が\SI{50}{\percent}以上変化する場合もある.
ワイヤ駆動ロボットの挙動をモデリングするためには, ワイヤの全長が大きく変化する場合のワイヤの張力や歪を計測することが必要である.
そこで本研究で製作するワイヤ試験機では, ワイヤの全長が動的に変化する場合のワイヤ張力を計測可能にする.
}%

\section{Design of Wire Testing machine} \label{sec:design}
\subsection{Overview} \label{subsec:design-overview}
\switchlanguage%
{%
In this chapter, a detailed description of the design of the wire testing machine developed in this study is provided.
\figref{figure:wiretester-design} shows an overview of the wire testing machine's configuration.

Based on the discussion in \secref{sec:characteristics}, the design objectives for the wire testing machine are defined as follows:
\begin{enumerate}
  \item Eliminate initial wire stretch
  \item Measure tension loss for various combinations of wires and pulleys of different diameters
  \item Measure wire tension when the total wire length dynamically changes
\end{enumerate}
To achieve these objectives, the wire testing machine developed in this study incorporates the following three systems:
\begin{enumerate}
  \item Multi-Pulley Pre-Stretching System
  \item Passive Pulley Transmission Efficiency Measurement System
  \item Variable-Length Wire Dynamics Measurement System
\end{enumerate}
The following sections provide detailed descriptions of these three systems.

\begin{figure}[t]
  \centering
  \includegraphics[width=1.0\columnwidth]{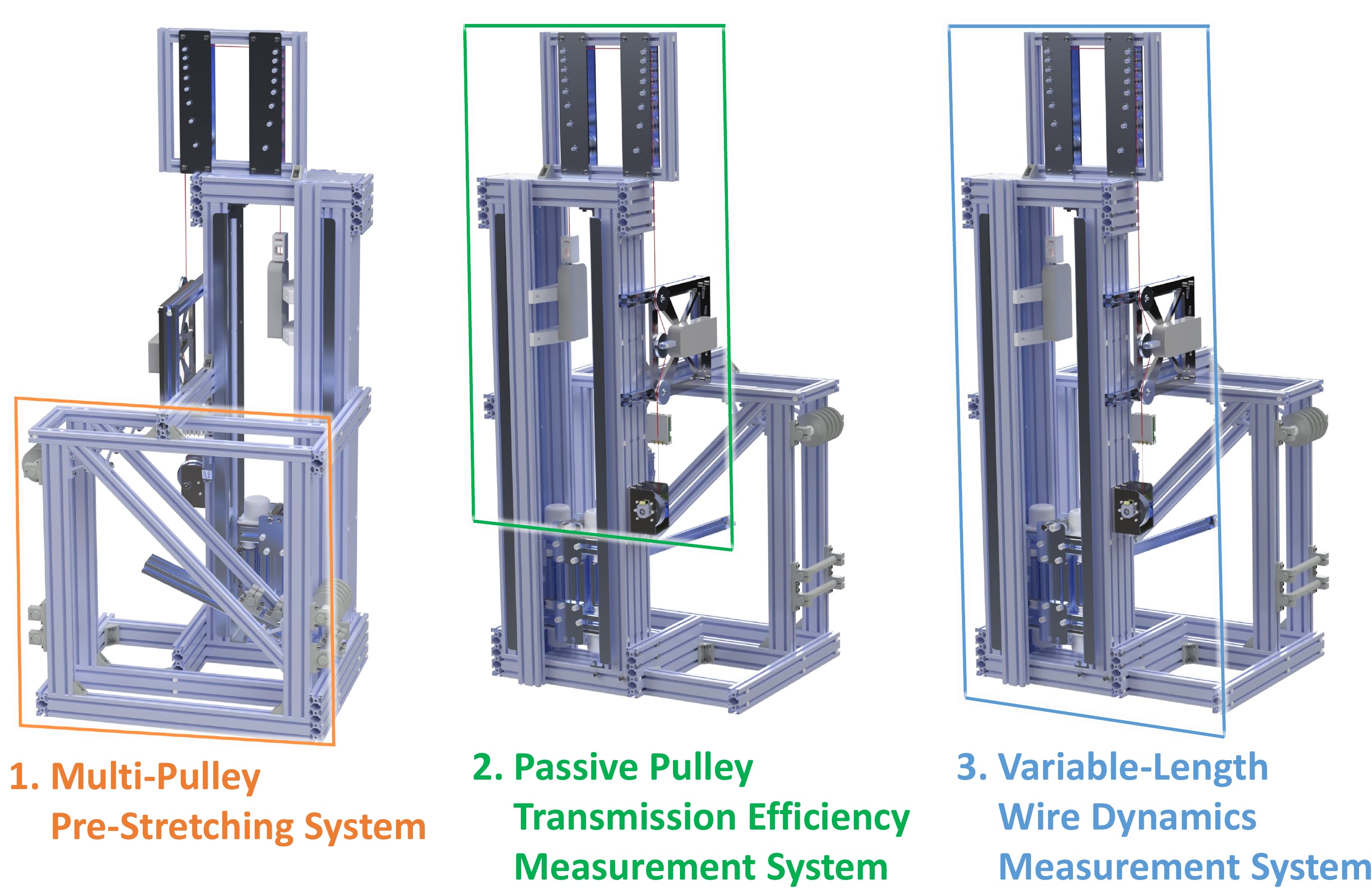}
  \vspace{-2ex}
  \caption{Design overview of wire testing machine.}
  \label{figure:wiretester-design}
  \vspace{-1.5ex}
\end{figure}
}%
{%
本章では, 製作したワイヤ試験機の設計について詳細な説明を行う.
\figref{figure:wiretester-design}にワイヤ試験機の設計の概要を示す.

\secref{sec:characteristics}での議論を元にワイヤ試験機の設計目標を以下の3つに定める.
\begin{enumerate}
  \item Eliminate initial wire stretch
  \item 多様なワイヤと多様な直径のプーリの組み合わせにおける張力損失の計測
  \item ワイヤの全長が動的に変化する場合のワイヤ張力の計測
\end{enumerate}
上記の目標を達成するために本研究で製作するワイヤ試験機に以下の3つの機構を搭載した.
\begin{enumerate}
  \item Multi-Pulley Pre-Stretching System
  \item Passive Pulley Transmission Efficiency Measurement System
  \item Variable-Length Wire Dynamics Measurement System
\end{enumerate}
以降の節ではこれら3つの機構について詳細な説明を行う.

\begin{figure}[t]
  \centering
  \includegraphics[width=1.0\columnwidth]{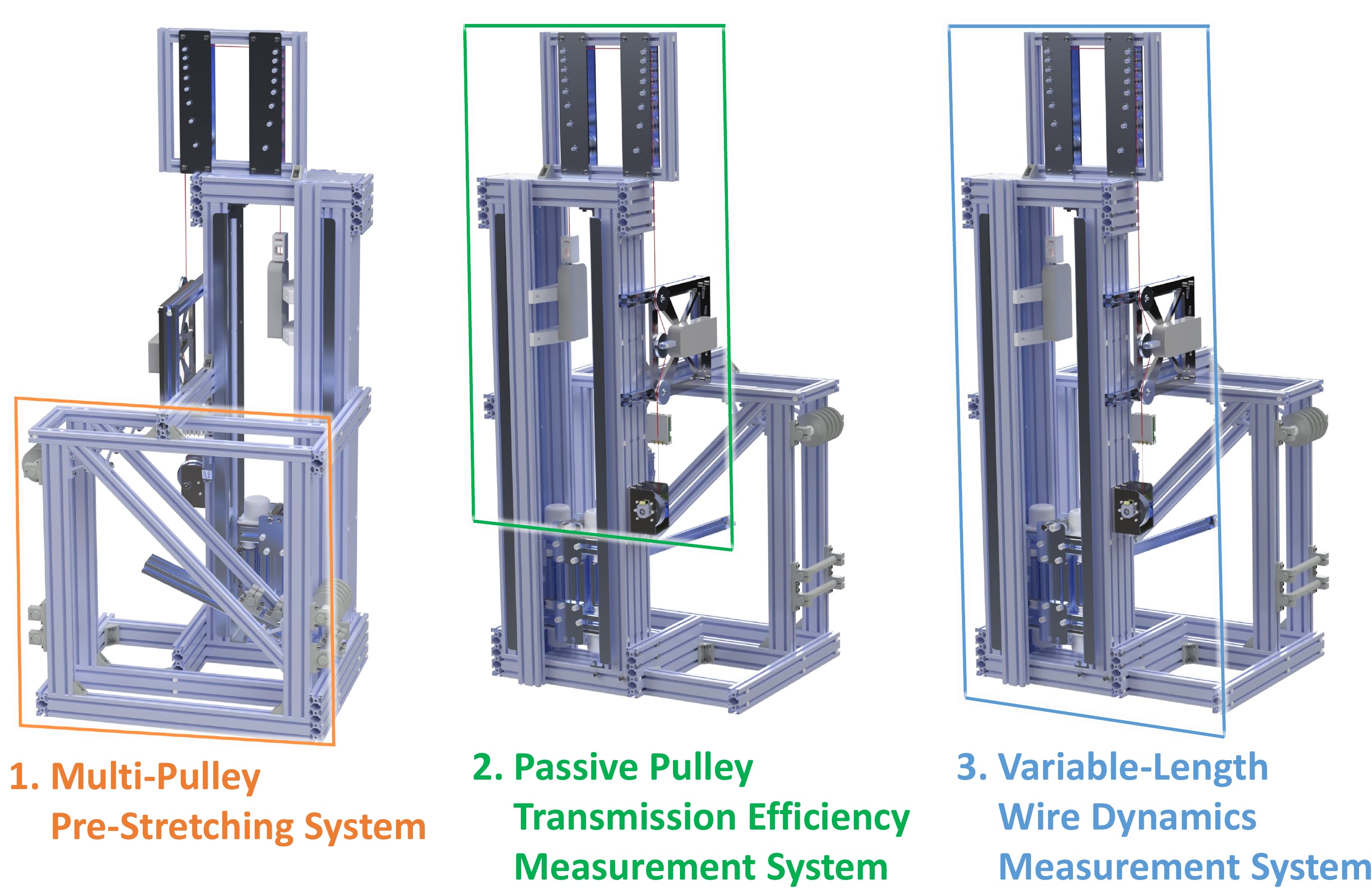}
  \vspace{-2ex}
  \caption{Design overview of wire testing machine.}
  \label{figure:wiretester-design}
  \vspace{-1.5ex}
\end{figure}
}%

\subsection{Multi-Pulley Pre-Stretching System} \label{subsec:wire-elongation}
\switchlanguage%
{%
The Multi-Pulley Pre-Stretching System applies tension to a new wire to eliminate initial wire stretch. 
The design of this system is shown in \figref{figure:elongation-design}.

This system comprises three components: a wire anchoring point, a multi-pulley for redirecting the wire, and a lever that applies tension. 
During setup, one end of the wire is first secured to the anchoring point, then routed over the multi-pulley, and finally attached to the lever. 
By placing a weight on the lever in this configuration, a constant tension is maintained on the wire, thereby removing initial wire stretch. 
Because the wire is looped five times by the multi-pulley, the mechanism remains compact (\SI{0.21}{\metre} x \SI{0.68}{\metre} x \SI{0.695}{\metre}) while enabling pre-stretching of up to approximately \SI{8.2}{\metre} of wire.

\figref{figure:elongation-test} shows the actual stretching of a \SI{3}{\milli\metre}-diameter Dyneema rope (DB-100, Hayami Industry). 
In this trial, a tension of \SI{510}{\newton} was applied to an \SI{8.2}{\metre} length of rope for 12 hours, resulting in a plastic deformation of \SI{0.43}{\metre}. 
These results confirm that the Multi-Pulley Pre-Stretching System can effectively eliminate initial wire stretch.

\begin{figure}[t]
  \centering
  \includegraphics[width=1.0\columnwidth]{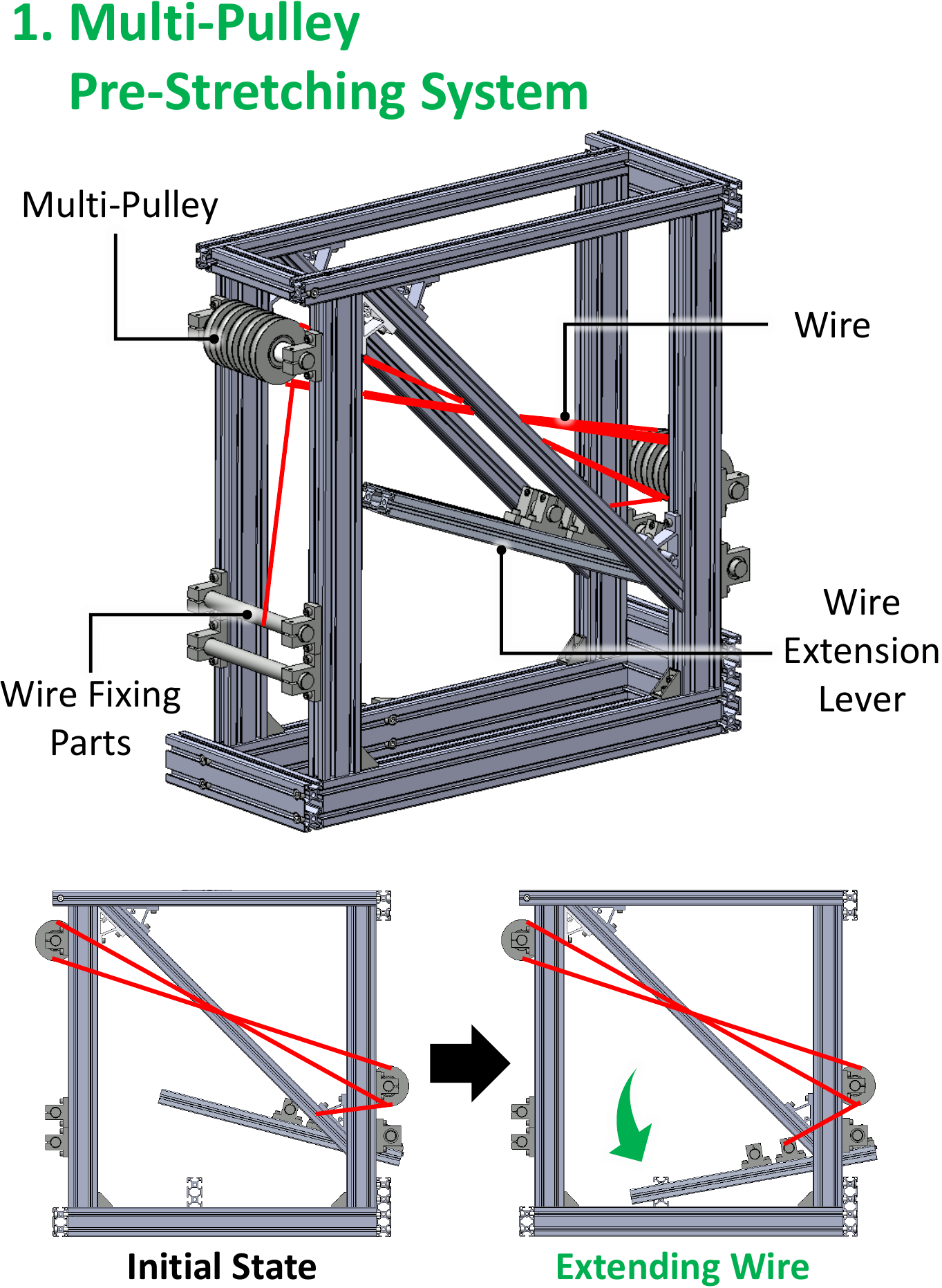}
  \vspace{-2ex}
  \caption{Detailed design of Multi-Pulley Pre-Stretching System.}
  \label{figure:elongation-design}
  \vspace{-0.5ex}
\end{figure}

\begin{figure}[t]
  \centering
  \includegraphics[width=1.0\columnwidth]{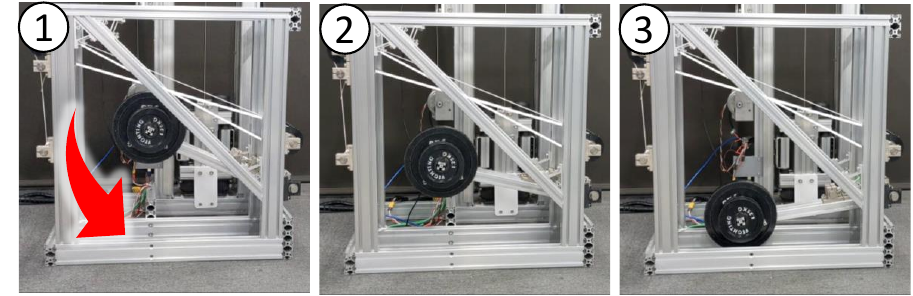}
  \vspace{-2ex}
  \caption{Snapshot of wire elongation test. The wire elongated from \SI{8.2}{\metre} to \SI{8.6}{\metre} in 12 hours.}
  \label{figure:elongation-test}
  \vspace{-1.5ex}
\end{figure}
}%
{%
Multi-Pulley Pre-Stretching Systemは新品のワイヤに張力を印加し, 初期伸びを除去する機構である.
本機構の設計を\figref{figure:elongation-design}に示す.

本機構はワイヤの固定端, ワイヤを折り返すためのmulti-pulley, ワイヤに張力を与えるレバーの3つの要素から成る.
ワイヤの取り付けでは, まず固定端に片端を取り付け, 次にmulti-pulleyに引っ掛け, 最後にレバーに逆端を固定する.
この状態でレバーに重りを搭載することでワイヤに一定張力を与え続け, 初期伸びを除去することができる.
multi-pulleyによってワイヤを5回折り返すため, \SI{0.21}{\metre} x \SI{0.68}{\metre} x \SI{0.695}{\metre}というコンパクトな機構でありながら最大約\SI{8.2}{\metre}のワイヤを延伸可能である.

\figref{figure:elongation-test}に実際に直径\SI{3}{\milli\metre}のDyneema rope (DB-100, Hayami industry)を伸ばしている図を示す.
本試行では, \SI{8.2}{\metre}のDyneema ropeに\SI{510}{\newton}の張力を12時間かけて, \SI{0.43}{\metre}の塑性変形を確認した.
本試行からMulti-Pulley Pre-Stretching Systemによって実際にワイヤの初期伸びを除去できることが確認できた.

\begin{figure}[t]
  \centering
  \includegraphics[width=1.0\columnwidth]{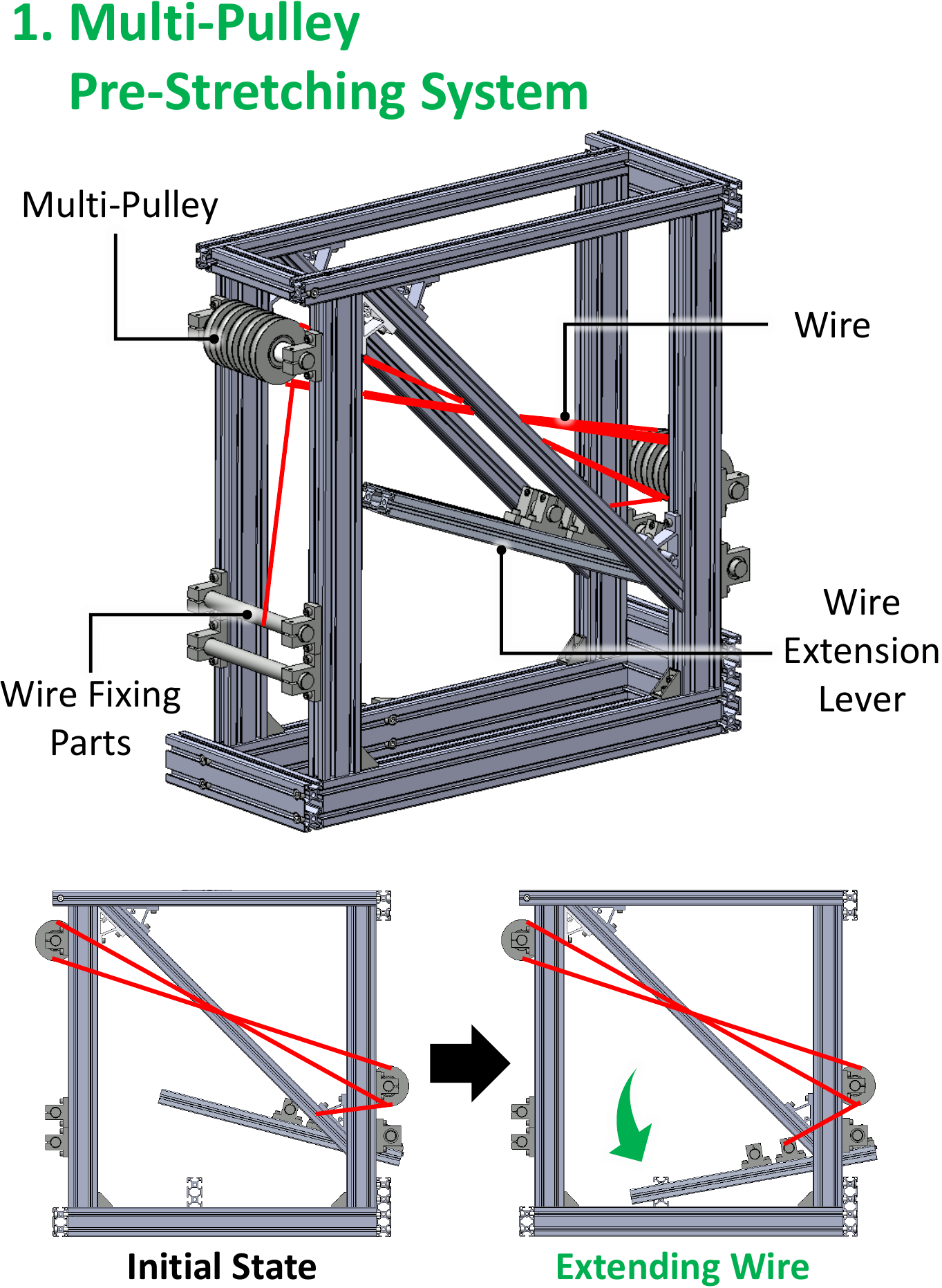}
  \vspace{-2ex}
  \caption{Detailed design of Multi-Pulley Pre-Stretching System.}
  \label{figure:elongation-design}
  \vspace{-1.5ex}
\end{figure}

\begin{figure}[t]
  \centering
  \includegraphics[width=1.0\columnwidth]{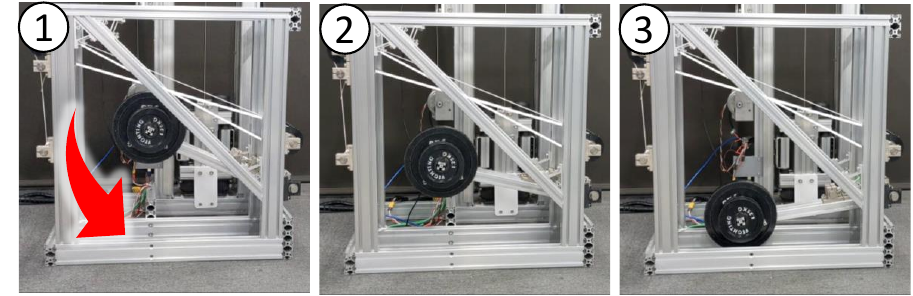}
  \vspace{-2ex}
  \caption{Snapshot of wire elongation test. The wire elongated from \SI{8.2}{\metre} to \SI{8.6}{\metre} in 12 hours.}
  \label{figure:elongation-test}
  \vspace{-1.5ex}
\end{figure}
}%

\subsection{Passive Pulley Transmission Efficiency Measurement System} \label{subsec:passive-pulley-measuremet}
\switchlanguage%
{%
The Passive Pulley Transmission Efficiency Measurement System measures tension transmission efficiency for combinations of wires with diameters from \SI{1}{\milli\metre} to \SI{3}{\milli\metre} and pulleys of eight diameters between \SI{12}{\milli\metre} and \SI{60}{\milli\metre}.
\figref{figure:passive-pulley-design} shows the design of this system.

This system comprises four components: a wire winding module, an input wire tension sensor, a passive pulley unit, and an output wire tension sensor. 
The wire winding module uses a motor (GL80 KV30, T-MOTOR) to reel the wire and can apply a maximum tension of \SI{400}{\newton}.

The input wire tension sensor measures the wire tension before the passive pulley using a force gauge (ZTS-500N, IMADA) and three passive pulleys (diameter \SI{40}{\milli\metre}). 
The input tension is calculated from the load on the force gauge and the corresponding geometric conditions.

The passive pulley unit houses two pulleys of each of eight diameters (\SI{12}{\milli\metre}, \SI{14}{\milli\metre}, \SI{16}{\milli\metre}, \SI{18}{\milli\metre}, \SI{20}{\milli\metre}, \SI{30}{\milli\metre}, \SI{40}{\milli\metre}, and \SI{60}{\milli\metre}). 
All pulleys are machined from 7075 aluminium alloy and anodized on the surface. 
Furthermore, each passive pulley incorporates a rolling bearing (F696ZZ, NSK).

The output wire tension sensor directly measures the wire tension after passing over the passive pulley using a force gauge (ZTS-500N, IMADA).

Whereas previous studies \cite{miyasaka2015measurement, choi2017tension, mate2024measurement} investigated tension transmission efficiency for only two pulley diameters, this apparatus can characterize transmission efficiency for eight diameters ranging from \SI{12}{\milli\metre} to \SI{60}{\milli\metre}, thereby supporting the design and control of a broader variety of wire-driven mechanisms.

\begin{figure}[t]
  \centering
  \includegraphics[width=1.0\columnwidth]{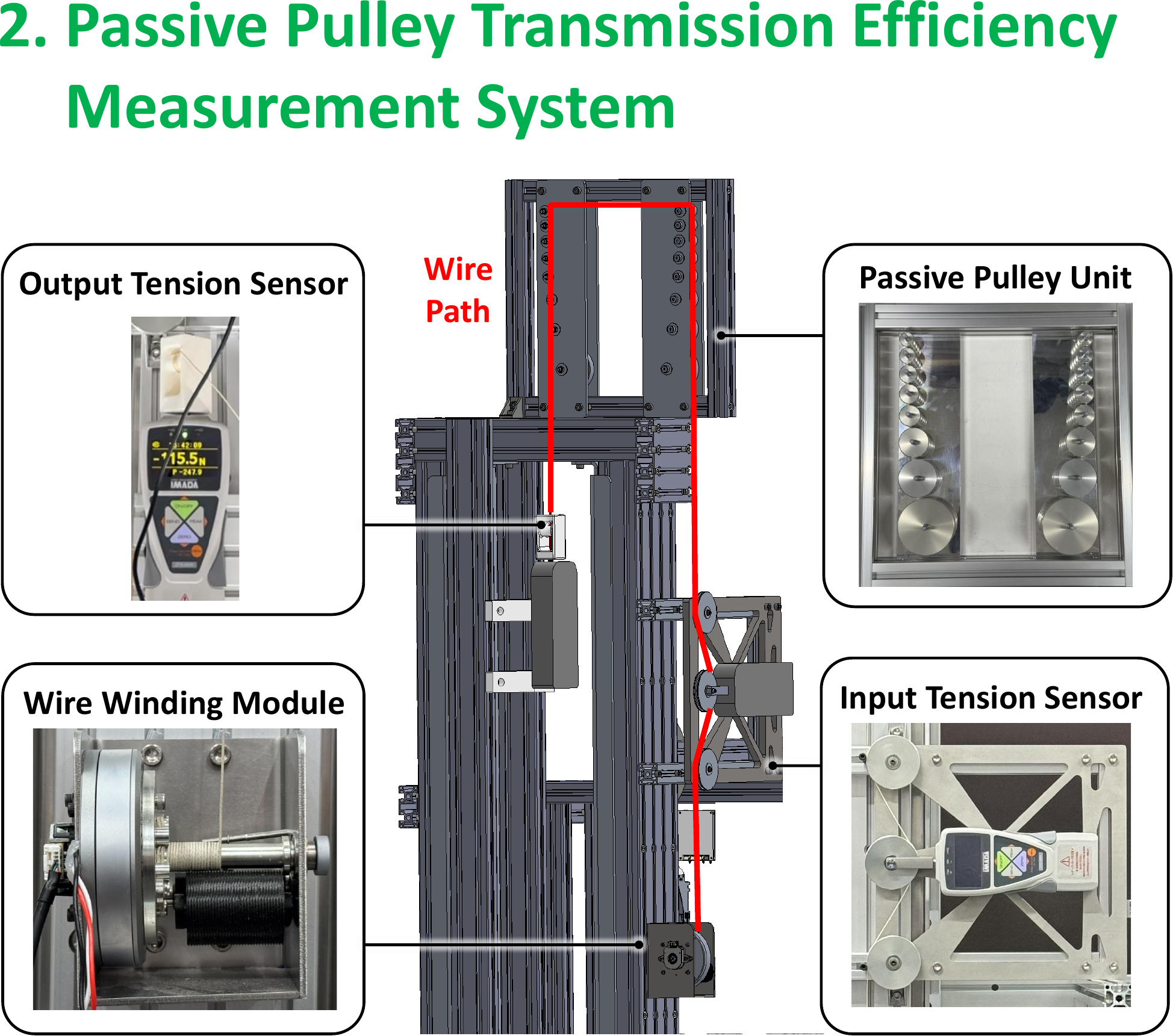}
  \vspace{-2ex}
  \caption{Detailed design of Passive Pulley Transmission Efficiency Measurement System.}
  \label{figure:passive-pulley-design}
  \vspace{-1.5ex}
\end{figure}
}%
{%
Passive Pulley Transmission Efficiency Measurement Systemは\SI{1}{\milli\metre} - \SI{3}{\milli\metre}の直径のワイヤと, \SI{12}{\milli\metre} - \SI{60}{\milli\metre}の8種類の直径のプーリの組み合わせにおける張力伝達効率を測定する機構である.
本機構の設計を\figref{figure:passive-pulley-design}に示す.

本機構はワイヤ巻取り装置, 入力ワイヤ張力測定装置, 受動プーリユニット, 出力ワイヤ張力測定装置の4つの要素から成る.
ワイヤ巻取り装置はモータ(GL80 KV30, T-MOTOR)でワイヤを巻き取る構造になっており, 最大400Nの張力を印加可能である.

入力ワイヤ張力測定装置は, 受動プーリを経由する前のワイヤ張力をフォーズゲージ(ZTS-500N, IMADA)と3つの受動プーリ(直径\SI{40}{\milli\metre})で測定する.
フォースゲージにかかる荷重から幾何的条件を利用して入力張力を測定する.

受動プーリユニットには8種類の直径(\SI{12}{\milli\metre}, \SI{14}{\milli\metre}, \SI{16}{\milli\metre}, \SI{18}{\milli\metre}, \SI{20}{\milli\metre}, \SI{30}{\milli\metre}, \SI{40}{\milli\metre}, \SI{60}{\milli\metre}) のプーリが各種2個ずつ搭載されている.
受動プーリはすべて超久ジュラルミン(A7075)製で, 表面にアルマイト処理を施してある.
さらにすべての受動プーリは転がり軸受(F696ZZ, NSK)を内蔵している.

出力ワイヤ張力測定装置は, 受動プーリ経由後のワイヤ張力をフォースゲージ(ZTS-500N, IMADA)で直接測定する.

先行研究\cite{miyasaka2015measurement, choi2017tension, mate2024measurement}では2種類程度の直径のプーリにおける張力伝達効率のみが調査されていたが, 本装置は\SI{12}{\milli\metre} - \SI{60}{\milli\metre}の8種類の直径のプーリの伝達効率を測定することができるため, より幅広いワイヤ駆動ロボットの設計・制御に役立つ.

% さらに, 先行研究\cite{choi2017tension}では直動機構でワイヤを引っ張ることで張力を印加していたのに対し, 本機構では回転機構(モータ)でワイヤを巻き取ることで張力を印加する構造になっている.
% 実際のワイヤ駆動ロボットの多く\cite{wamarm2024, kim2017anthropomorphic, song2018lims2, suzuki2024saqiel, kawaharazuka2019musashi, endo2019superdragon}は回転機構(モータ)を用いてワイヤを駆動するため, 本試験機を用いることでより実際のワイヤ駆動ロボットと近い条件で伝達効率を測定できる.
\begin{figure}[t]
  \centering
  \includegraphics[width=1.0\columnwidth]{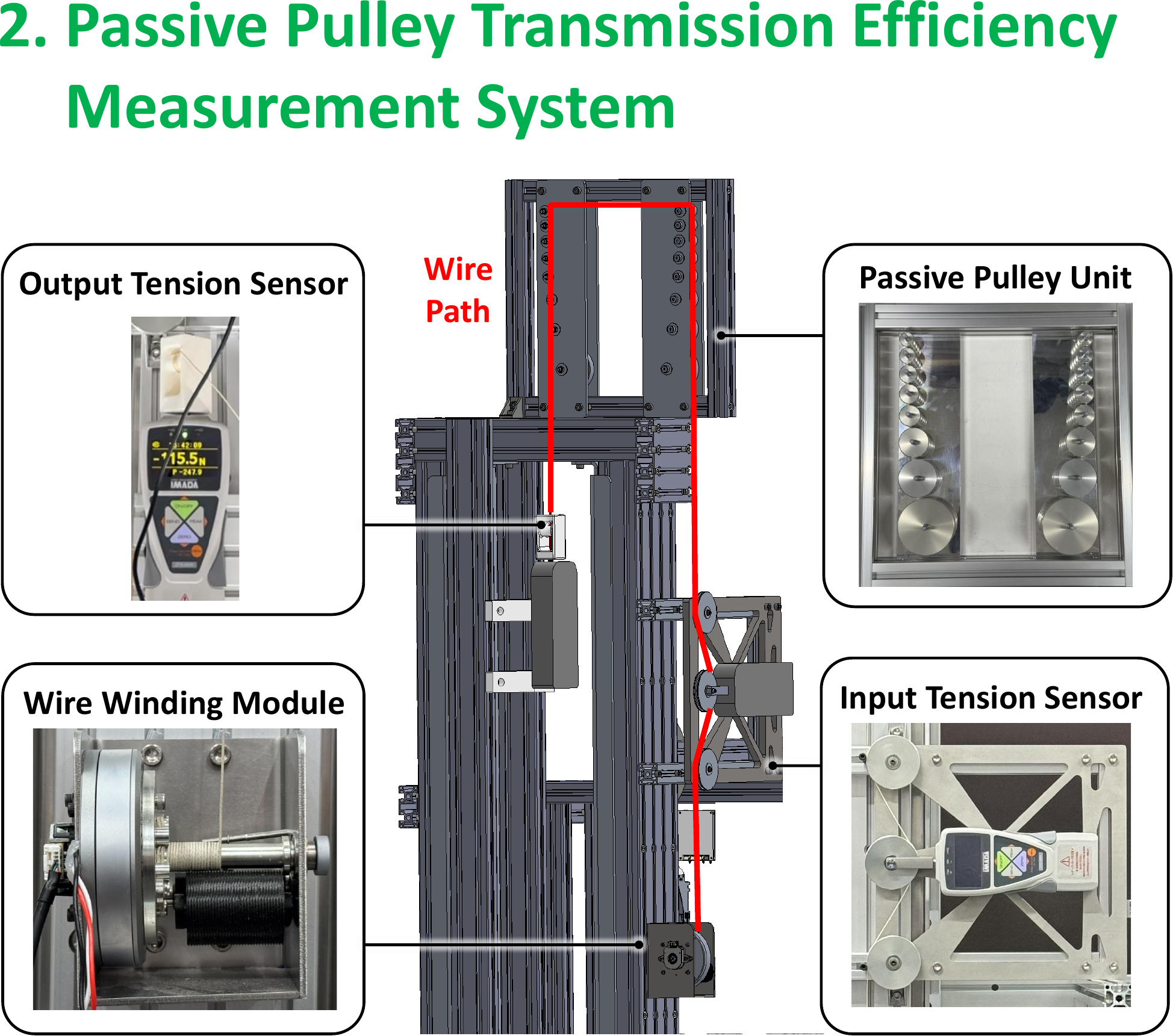}
  \vspace{-2ex}
  \caption{Detailed design of Passive Pulley Transmission Efficiency Measurement System.}
  \label{figure:passive-pulley-design}
  \vspace{-1.5ex}
\end{figure}
}%

\subsection{Variable-Length Wire Dynamics Measurement System} \label{subsec:linear-load-testing}
\switchlanguage%
{%
The Variable-Length Wire Dynamics Measurement System is an apparatus that measures wire tension and strain when a load is moved up and down by a wire. 
The design of this system is shown in \figref{figure:linear-load-design}.

This system consists of four components: a wire winding module, an input tension sensor, a passive pulley unit, and a linear loading unit. 
Since this apparatus replaces the output tension sensor of the Passive Pulley Transmission Efficiency Measurement System with the linear loading unit, the three components other than the linear loading unit are described in \subsecref{subsec:passive-pulley-measuremet}. 
The linear loading unit secures an \SI{8.1}{\kilogram} mass on a linear guide and moves it ±\SI{0.35}{\metre} vertically. Load displacement is measured by a linear encoder (LF-11, RLS). 
When the wire winding module reels in the wire, the load rises via the wire attached to the top of the linear loading unit.

Using this apparatus, it is possible to measure the dynamics under conditions in which the total wire length changes, as occurs in actual wire-driven robots.

\begin{figure}[t]
  \centering
  \includegraphics[width=1.0\columnwidth]{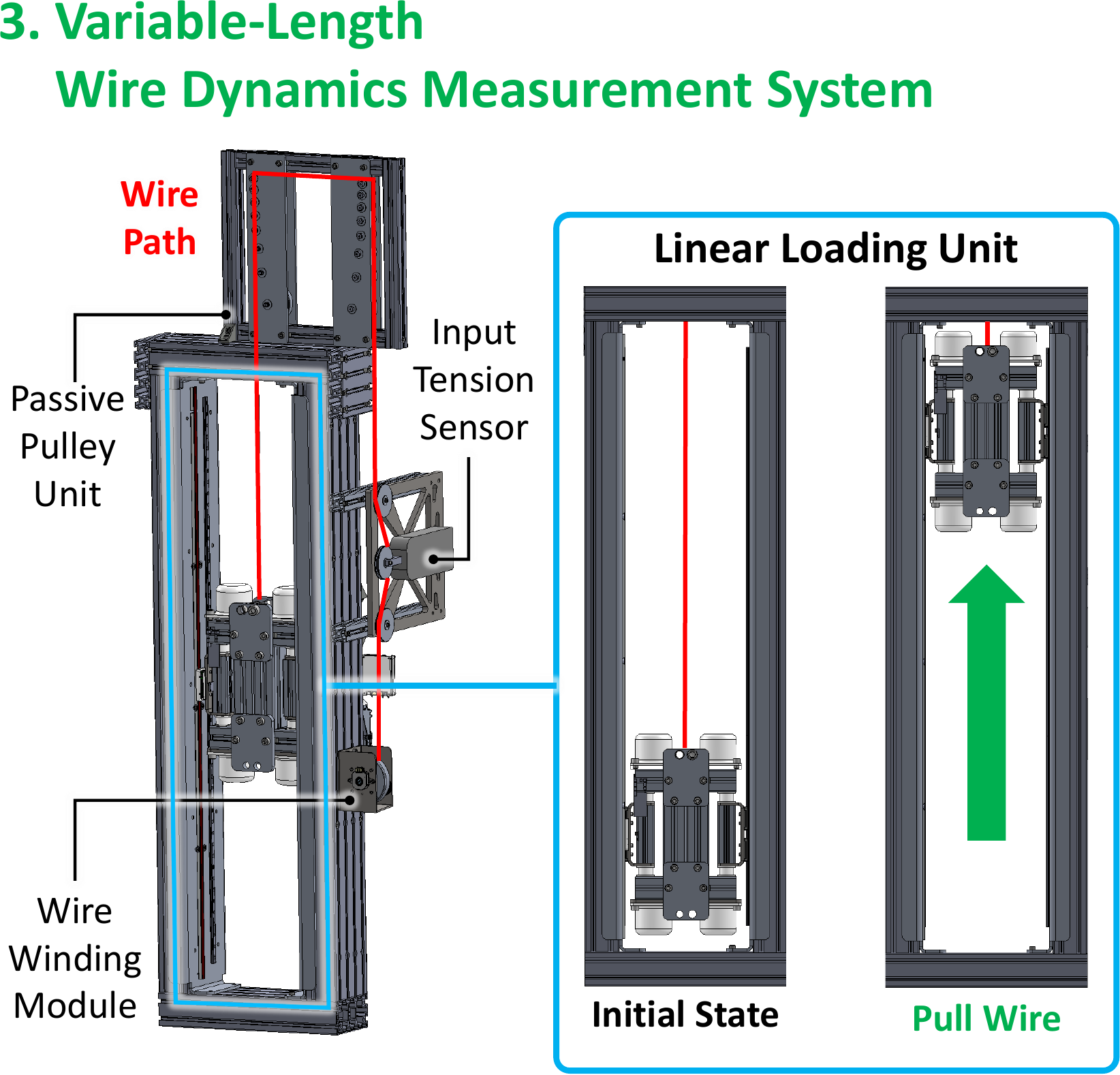}
  \vspace{-2ex}
  \caption{Detailed design of Variable-Length Wire Dynamics Measurement System.}
  \label{figure:linear-load-design}
  \vspace{-1.5ex}
\end{figure}
}%
{%
Variable-Length Wire Dynamics Measurement Systemは, ワイヤを用いて負荷を上下させたときの張力及び歪を測定できる装置である.
本機構の設計を\figref{figure:linear-load-design}に示す.

本機構はワイヤ巻取り装置, 入力ワイヤ張力測定装置, 受動プーリユニット, 直動負荷装置の4つの要素から成る.
本装置はPassive Pulley Transmission Efficiency Measurement Systemの出力張力測定装置を直動負荷に換装したものであるため, 直動負荷装置以外の3要素は\subsecref{subsec:passive-pulley-measuremet}を参照.
直動負荷装置は\SI{8.1}{\kilogram}の重りがリニアガイド上に固定されており±\SI{0.35}{\metre}上下する.
負荷の変位はリニアエンコーダ(LF-11, RLS)で測定できる.
ワイヤ巻取り装置でワイヤを引っ張ると, 直動負荷装置の上部に取り付けられたワイヤによって負荷が上昇する. 

本装置を用いることで, 実際のワイヤ駆動ロボットのようにワイヤの全長が変化する条件でのdynamicsを測定できる.

\begin{figure}[t]
  \centering
  \includegraphics[width=1.0\columnwidth]{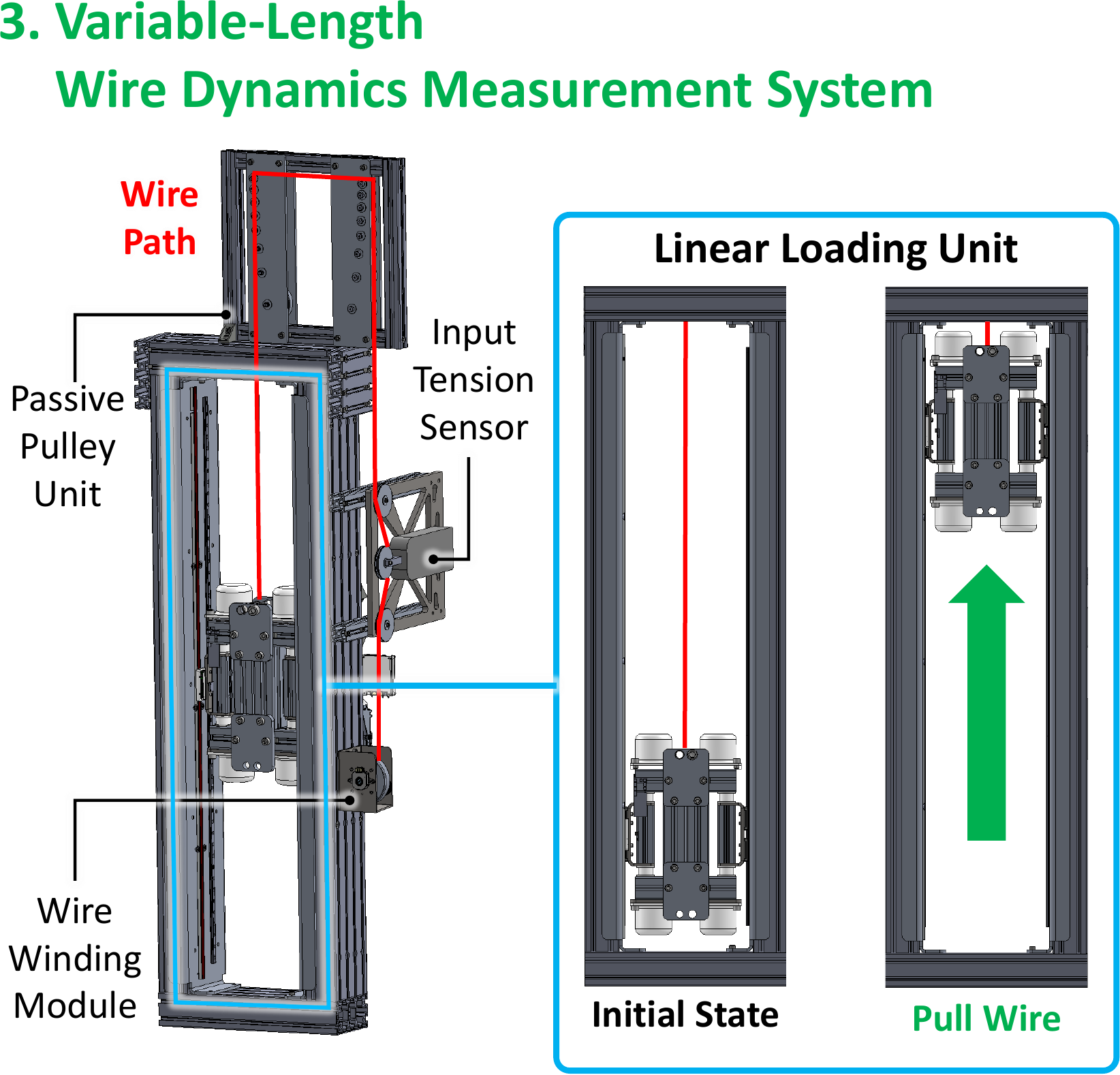}
  \vspace{-2ex}
  \caption{Detailed design of Variable-Length Wire Dynamics Measurement System.}
  \label{figure:linear-load-design}
  \vspace{-1.5ex}
\end{figure}
}%

\section{Experiments} \label{sec:experiments}
\switchlanguage%
{%
In this chapter, to evaluate each function of the wire testing machine, we measure the tension transmission efficiency of the passive pulley and the dynamic characteristics of the variable-length wire. 
We then apply the measured data to the force control of a wire-driven robot to verify the usefulness of the wire testing machine.
}%
{%
本章ではワイヤ試験機の各機能を検証するため, 受動プーリにおける張力伝達効率の測定, 可変長ワイヤの動特性の測定を行う. 
そして測定したデータをワイヤ駆動ロボットの力制御に適用し, ワイヤ試験機の有用性を検証する.
}

\subsection{Measurement of Tension Transmission Efficiency in Wire Pulleys} \label{subsec:pulley-efficiency}
\switchlanguage%
{%
Using the Passive Pulley Transmission Efficiency Measurement System, we measured the tension transmission efficiency for combinations of eight pulley diameters and four wire types. 
The pulleys had diameters of \SI{12}{\milli\metre}, \SI{14}{\milli\metre}, \SI{16}{\milli\metre}, \SI{18}{\milli\metre}, \SI{20}{\milli\metre}, \SI{30}{\milli\metre}, \SI{40}{\milli\metre}, and \SI{60}{\milli\metre}. 
The wires comprised a \SI{1}{\milli\metre}-diameter Vectran rope (VB-175, Hayami industry) and Zylon ropes of \SI{2}{\milli\metre}, \SI{2.5}{\milli\metre}, and \SI{3}{\milli\metre} diameters (SZ-20, SZ-25, SZ-30, Hayami industry). 
We investigated the tension transmission efficiency under applied tensions of \SI{200}{\newton} and \SI{400}{\newton}. 
For each pulley, wire, and applied tension, twenty measurements were taken and the average value of the tension transmission efficiency was calculated.

The transmission efficiency in the passive pulley was determined by measuring the tension before and after passing over the pulley ($T_{\mathrm{in}}$ and $T_{\mathrm{out}}$).
Since the apparatus routes the wire over two passive pulleys, the transmission efficiency per passive pulley $E$ was calculated as
$E = \sqrt{T_{\mathrm{out}} / T_{\mathrm{in}}}$.

The experimental results under tensions of \SI{200}{\newton} and \SI{400}{\newton} are shown in \figref{figure:efficiency-200N} and \figref{figure:efficiency-400N}, respectively.
For the smallest pulley with a diameter of \SI{12}{\milli\metre}, the ratio of lost tension to input tension was 0.021-0.081.
This value is significantly larger than the typical friction coefficient of deep-groove ball bearings (0.001-0.0015)\cite{bearing-jtekt}.
These findings confirm, as pointed out in prior work\cite{mate2024measurement}, that the primary source of tension loss when passing through a passive pulley is not bearing friction but fiber-fiber interaction within the wire or friction between the wire and the pulley surface.

For all wire types and applied tensions, it was confirmed that the tension transmission efficiency increases monotonically with pulley diameter.
Moreover, for a given pulley diameter, the tension transmission efficiency decreases as the wire diameter increases.
These experiments demonstrate that by increasing the passive pulley diameter and reducing the wire diameter, a low-friction transmission mechanism can be realized in wire-driven robots.

\begin{figure}[t]
  \centering
  \includegraphics[width=1.0\columnwidth]{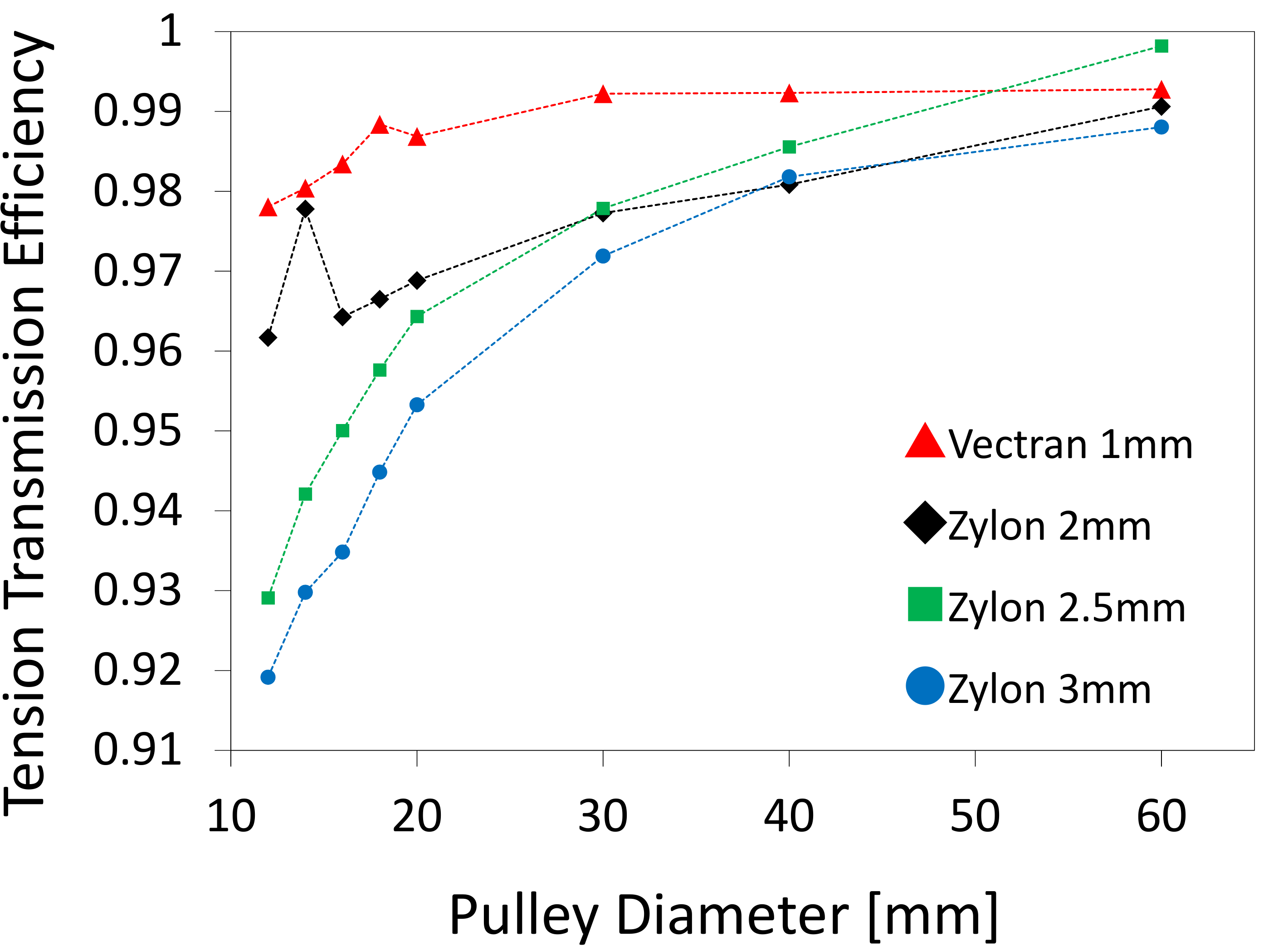}
  \vspace{-2ex}
  \caption{The relationship between pulley diameter and tension transmission efficiency at 200N tension force.}
  \label{figure:efficiency-200N}
  \vspace{-0.5ex}
\end{figure}

\begin{figure}[t]
  \centering
  \includegraphics[width=1.0\columnwidth]{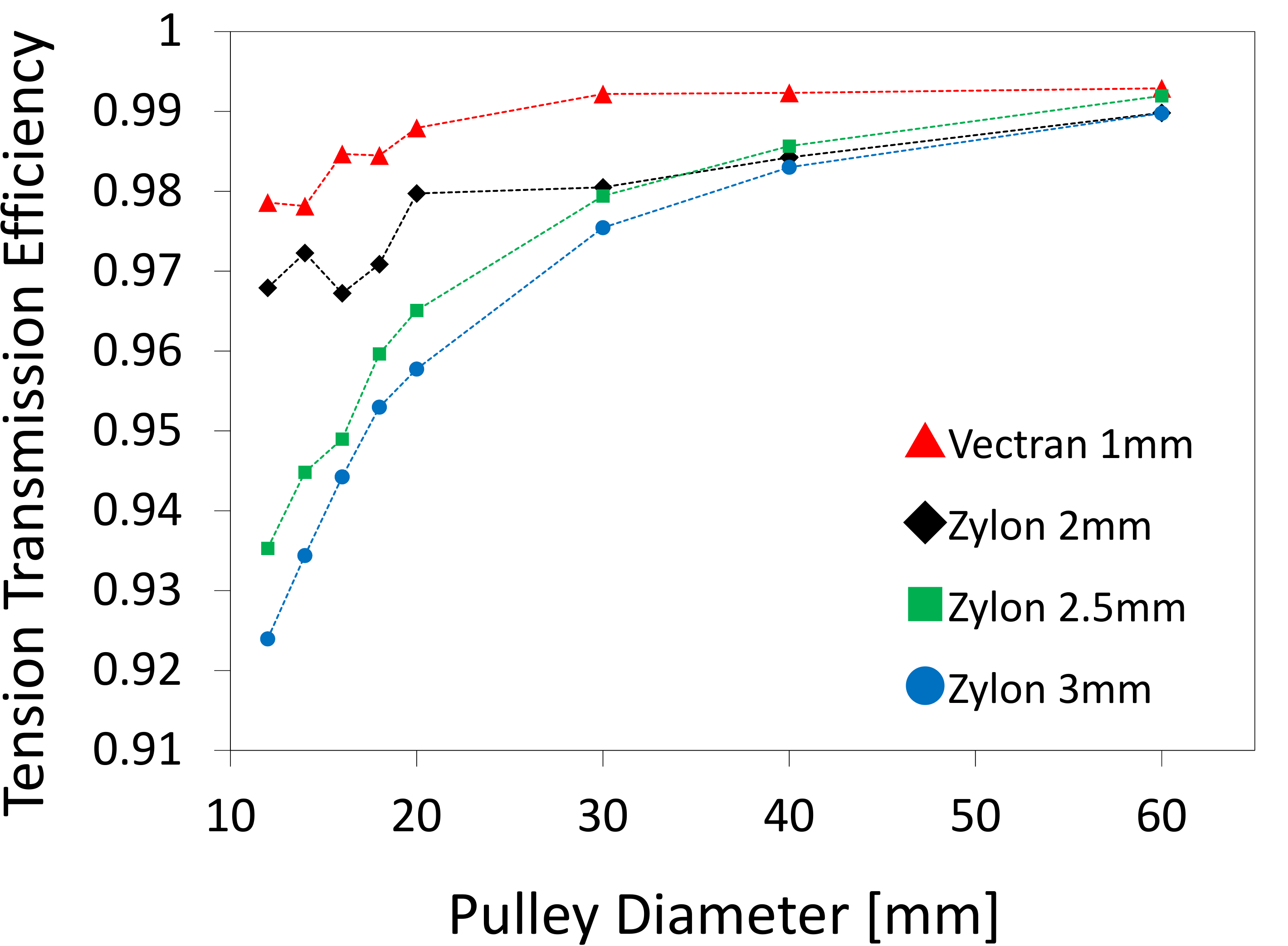}
  \vspace{-2ex}
  \caption{The relationship between pulley diameter and tension transmission efficiency at 400N tension force.}
  \label{figure:efficiency-400N}
  \vspace{-1.5ex}
\end{figure}
}%
{%
Passive Pulley Transmission Efficiency Measurement Systemを用いて, 8種類の直径のプーリと4種類のワイヤの組み合わせにおける張力伝達効率を測定した.
プーリは直径\SI{12}{\milli\metre}, \SI{14}{\milli\metre}, \SI{16}{\milli\metre}, \SI{18}{\milli\metre}, \SI{20}{\milli\metre}, \SI{30}{\milli\metre}, \SI{40}{\milli\metre}, \SI{60}{\milli\metre}の8種類を使用した.
ワイヤは直径\SI{1}{\milli\metre}のVectran rope (VB-175, Hayami industry)と, 直径\SI{2}{\milli\metre}, \SI{2.5}{\milli\metre}, \SI{3}{\milli\metre}のZylon rope (SZ-20, SZ-25, SZ-30, Hayami industry)の4種類を使用した.
そして\SI{200}{\newton}と\SI{400}{\newton}の2種類の張力を印加した場合の伝達効率を調べた.
なお各プーリ, ワイヤ, 印加張力においてそれぞれ20回ずつ伝達効率を測定し, その平均値を計算した.

受動プーリにおける伝達効率はプーリ経由前の張力$T_{\mathrm{in}}$と経由後の張力$T_{\mathrm{out}}$を測定することで求めた.
本装置では2つの受動プーリを経由するため, 受動プーリ1個あたりの伝達効率$E$は$E = \sqrt{T_{\mathrm{out}} / T_{\mathrm{in}}}$と計算した.

張力\SI{200}{\newton}と\SI{400}{\newton}における実験結果をそれぞれ\figref{figure:efficiency-200N}と\figref{figure:efficiency-400N}に示す.
最も小さい直径\SI{12}{\milli\metre}のプーリでは, 入力張力に対する損失張力の割合は0.021-0.081である.
これは一般的な深溝玉軸受の摩擦係数0.001-0.0015\cite{bearing-jtekt}に対して優位に大きい.
このことから先行研究\cite{mate2024measurement}で指摘されていた通り, 受動プーリを経由する際の張力損失の主要因はベアリング内部の摩擦ではなく, ワイヤ内部の繊維同士の擦れまたは, ワイヤとプーリ表面の摩擦であることが確認できる.

すべてのワイヤと印加張力において, 張力伝達効率がプーリ直径とともに単調増加していることが確認できる.
またプーリ直径が等しい場合は, ワイヤ直径が大きいほど張力伝達効率が小さくなっている.
本実験から, 受動プーリの直径を大きくしワイヤの直径を小さくすることでワイヤ駆動ロボットの駆動系の低摩擦化が実現できることが分かる.

\begin{figure}[t]
  \centering
  \includegraphics[width=1.0\columnwidth]{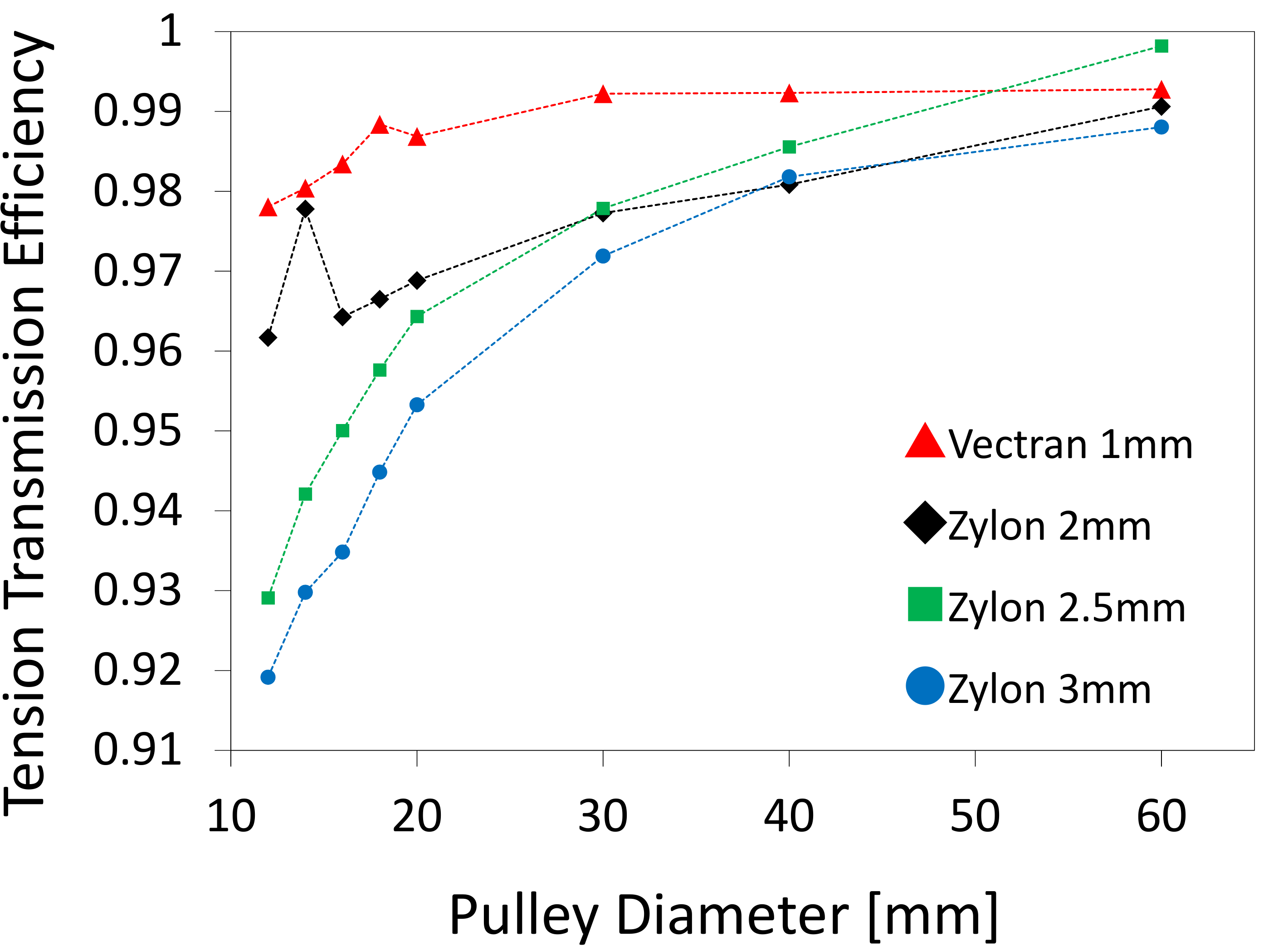}
  \vspace{-2ex}
  \caption{The relationship between pulley diameter and tension transmission efficiency at 200N tension force.}
  \label{figure:efficiency-200N}
  \vspace{-1.5ex}
\end{figure}

\begin{figure}[t]
  \centering
  \includegraphics[width=1.0\columnwidth]{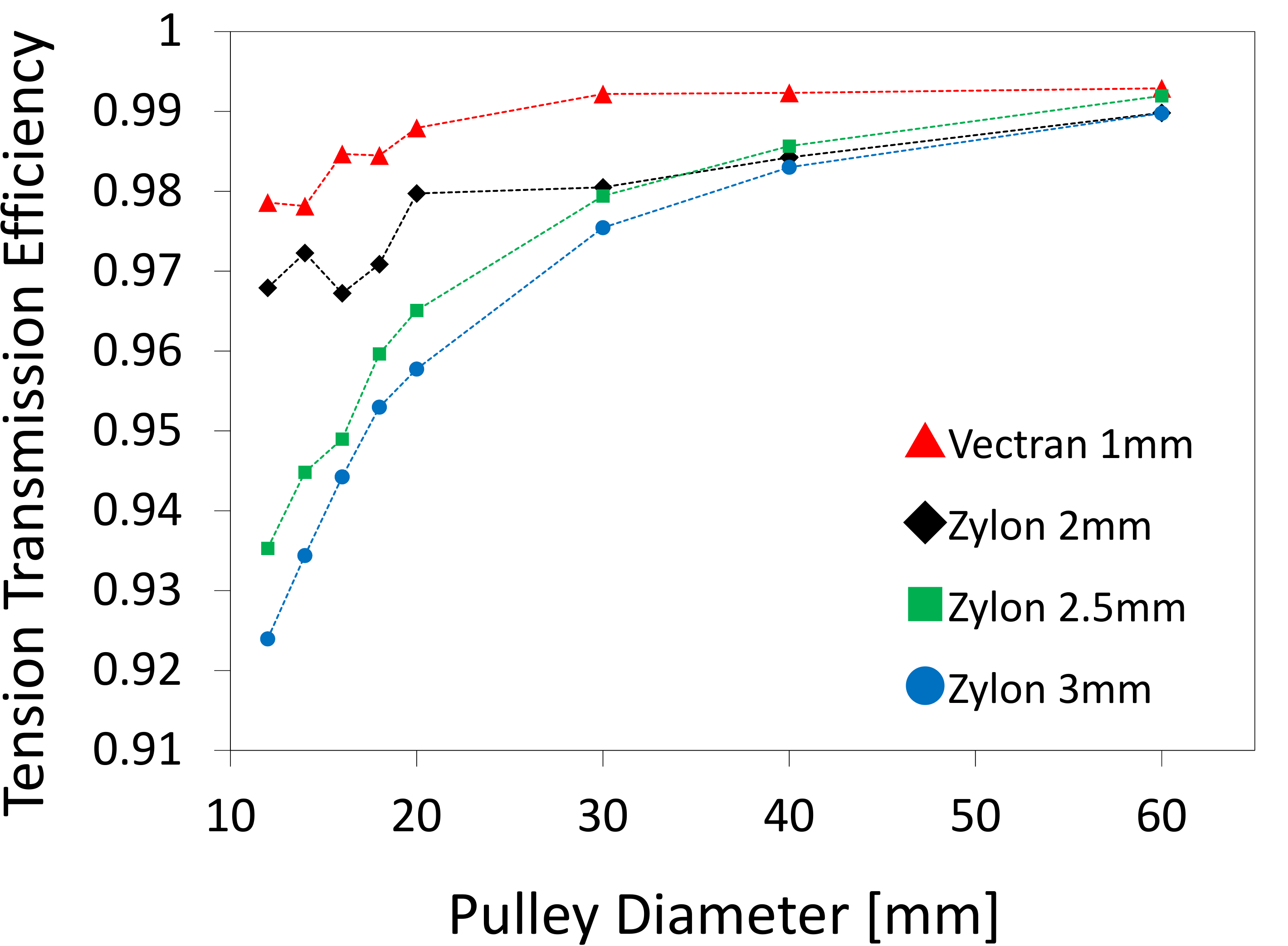}
  \vspace{-2ex}
  \caption{The relationship between pulley diameter and tension transmission efficiency at 400N tension force.}
  \label{figure:efficiency-400N}
  \vspace{-1.5ex}
\end{figure}
}%

\subsection{Measurement of Dynamic Characteristics of Wires} \label{subsec:dynamic-experiment}
\switchlanguage%
{%
Using the Variable-Length Wire Dynamics Measurement System, we investigated the frequency response of wire tension when the wire length varied and when it remained constant.
In this experiment, a sinusoidal wave with constant amplitude and time-varying frequency was input as the tension command to the wire testing machine, and the actual tension was measured.
The tension command had a minimum of \SI{0}{\newton}, a maximum of \SI{150}{\newton}, a lowest frequency of \SI{2}{\hertz}, and a highest frequency of \SI{20}{\hertz}.
We examined the frequency response between the commanded tension and the actual tension in two conditions: with the linear loading unit free to move and with its lower end fixed.
In this experiment, a \SI{1}{\milli\metre}-diameter Vectran rope (VB-175, Hayami industry) was used as the driving wire.

\figref{figure:frequency-response} shows the frequency response of tension for the variable-length wire and the constant-length wire.
The magnitude of the variable-length wire is approximately \SI{-5}{\decibel} at \SI{2}{\hertz}, and a small resonance peak of about \SI{+3}{\decibel} appears near \SI{9}{\hertz}.
The magnitude of the constant-length wire remains around \SI{0}{\decibel} in the low-frequency range, while showing a sharp resonance peak of approximately \SI{+6}{\decibel} near \SI{6}{\hertz}.
The phase lag of the variable-length wire remains below \SI{-30}{\degree} between \SI{2}{\hertz} and \SI{8}{\hertz}, and progresses rapidly to \SI{-160}{\degree} between \SI{8}{\hertz} and \SI{10}{\hertz}.
The phase lag of the constant-length wire progresses rapidly to \SI{-160}{\degree} between \SI{4}{\hertz} and \SI{10}{\hertz}.
From these results, the dynamic change in wire length reduces tension tracking performance by approximately \SI{5}{\decibel} at low frequencies, while halving the resonance amplification in the mid-to-high frequency range and mitigating the progression of phase lag.

This experiment revealed a significant difference in tension characteristics between the variable-length wire and the constant-length wire, demonstrating the importance of the Variable-Length Wire Dynamics Measurement System.

\begin{figure}[t]
  \centering
  \includegraphics[width=1.0\columnwidth]{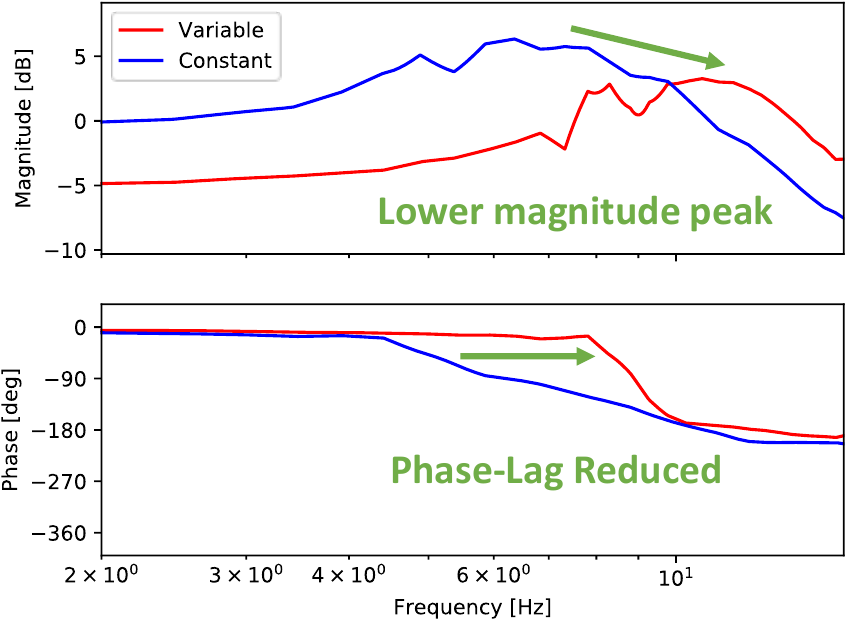}
  \vspace{-2ex}
  \caption{Frequency response of tension for constant and variable total wire length.}
  \label{figure:frequency-response}
  \vspace{-1.5ex}
\end{figure}
}%
{%
Variable-Length Wire Dynamics Measurement Systemを用いて, ワイヤの全長が変化する場合と一定の場合のワイヤ張力の周波数応答を調べた.
本実験ではワイヤ試験機の張力指令値として振幅一定で周波数が時間とともに増加するsin波を入力し, 実際の張力を測定した.
張力指令値は最小値\SI{0}{\newton}, 最大値\SI{150}{\newton}, 最低周波数\SI{2}{\hertz}, 最高周波数\SI{20}{\hertz}である.
ワイヤが結ばれた直動ユニットを自由に運動させた状態と, 下端に固定した状態で目標張力と実張力の間の周波数応答をそれぞれ調べた.
なお本実験では駆動用ワイヤとして直径\SI{1}{\milli\metre}のVectran rope (VB-175, Hayami industry)を用いた.

\figref{figure:frequency-response}に可変長ワイヤと固定長ワイヤの張力の周波数応答を示す.
可変長ワイヤのmagnitudeは\SI{2}{\hertz}で約\SI{-5}{\decibel}であり、\SI{9}{\hertz}付近で約\SI{+3}{\decibel}の小さな共振ピークが現れている。
固定長ワイヤのmagnitudeは低周波数帯で約\SI{0}{\decibel}を維持する一方, \SI{6}{\hertz}付近で約\SI{+6}{\decibel}の鋭い共振ピークを示している。
可変長ワイヤの位相遅れは\SI{2}{\hertz}-\SI{8}{\hertz}で\SI{-30}{\degree}以下を維持し, \SI{8}{\hertz}-\SI{10}{\hertz}で\SI{-160}{\degree}まで急激に進行している.
固定長ワイヤの位相遅れは\SI{4}{\hertz}-\SI{10}{\hertz}で\SI{-160}{\degree}まで急激に進行している.
以上より、ワイヤ全長の動的変化は低周波での張力追従性を約\SI{5}{\decibel}低減させるものの, 中高周波の共振増幅を半減させ, 位相遅れの進行を緩和させることが分かった.

本実験から可変長ワイヤと固定長ワイヤの張力特性が大きく異なることが明らかになり, Variable-Length Wire Dynamics Measurement Systemの重要性が示された.

\begin{figure}[t]
  \centering
  \includegraphics[width=1.0\columnwidth]{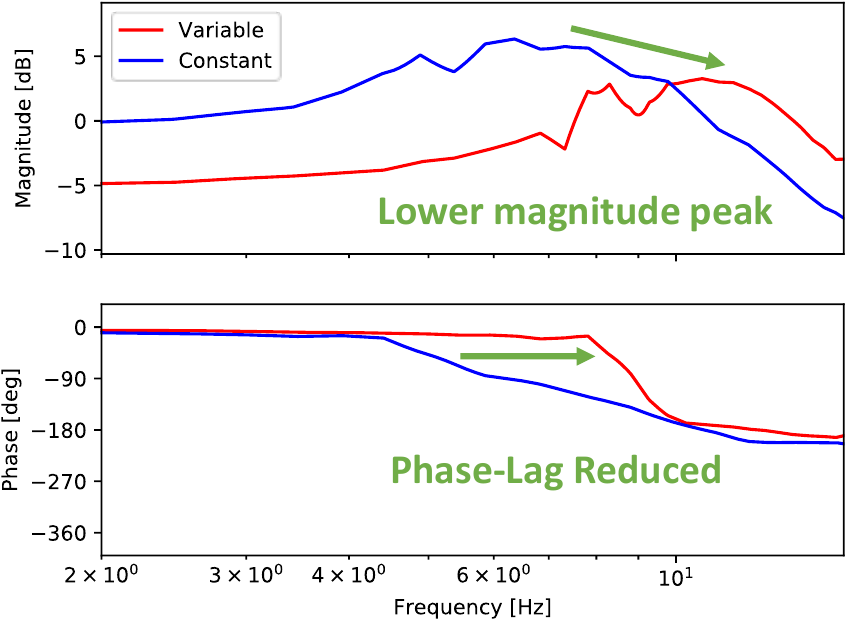}
  \vspace{-2ex}
  \caption{Frequency response of tension for constant and variable total wire length.}
  \label{figure:frequency-response}
  \vspace{-1.5ex}
\end{figure}
}%

\subsection{Wire-Driven Manipulator Control Considering Tension Transmission Efficiency in Wire Pulleys} \label{subsec:saqiel-control}
\switchlanguage%
{%
Using the tension transmission efficiency of the passive pulley measured in \subsecref{subsec:pulley-efficiency}, we applied it to the force control of an actual wire-driven robot.
The experimental setup is shown in \figref{figure:ee-force-setup}.
In this experiment, a target end-effector force was commanded to the coupled wire-driven mechanism, and the error between the commanded end-effector force and the actual force was measured by a six-axis force sensor at the end effector.
We then compared the end-effector force error with and without compensation for tension loss in the passive pulley.
The robot used in this experiment employed a \SI{1}{\milli\metre}-diameter Vectran rope (VB-175, Hayami industry) and \SI{12}{\milli\metre}-diameter passive pulleys.
In this robot, the wire passes through up to seven passive pulley units.

The end-effector force control law used in this experiment is described as follows.
First, the target end-effector force $\bm{f}{\mathrm{ref}}$ is provided by a higher-level controller.
Next, the target joint torque $\bm{\tau}{\mathrm{ref}}$ is computed from the target end-effector force $\bm{f}{\mathrm{ref}}$ and the joint Jacobian $\bm{J}{\mathrm{r}}$, which relates the robot’s end-effector position to its joint angles, as
\begin{align}
  \label{eq:eeforce-torque}
  \bm{\tau}_{\mathrm{ref}} = \bm{J}_{\mathrm{r}}^{\mathrm{T}} \bm{f}_{\mathrm{ref}}
\end{align}

From here, we explain the conversion from the target joint torque $\bm{\tau}{\mathrm{ref}}$ to the target wire tension $\bm{T}{\mathrm{ref}}$.
First, we compute the target wire tension $\bm{T}{\mathrm{ref}}$ without considering tension loss in the passive pulley.
The following equality holds between the target joint torque $\bm{\tau}{\mathrm{ref}}$ and the target wire tension $\bm{T}{\mathrm{ref}}$:
\begin{align}
  \label{eq:tension-torque}
  \bm{\tau}_{\mathrm{ref}} = -\bm{G}^{\mathrm{T}} \bm{T}_{\mathrm{ref}}
\end{align}
Here, $\bm{G}$ is the muscle length Jacobian matrix, which represents the moment arms of each wire at each joint.
The muscle length Jacobian $\bm{G}(\bm{q})$ at a given joint angle $\bm{q}$ is defined as
\begin{align}
  \label{eq:muscle-jacobian}
  \bm{G}(\bm{q}) = \frac{\partial \bm{l}}{\partial \bm{q}}
\end{align}
where $\bm{l}$ is the wire length.
Because the actual wire-driven robot is subject to limits on producible tension, the target wire tension $\bm{T}{\mathrm{ref}}$ (without considering tension loss in the passive pulley) is obtained by solving the following quadratic program \cite{kawamura2016jointspace}.
\begin{align}
  \label{calc-tension}
  \begin{split}
    \text{minimize} & \quad (\bm{\tau}_{\mathrm{ref}}+\bm{G}^{T}\bm{T}_{\mathrm{ref}})^{T} \Lambda (\bm{\tau}_{\mathrm{ref}}+\bm{G}^{T}\bm{T}_{\mathrm{ref}})+|\bm{T}_{\mathrm{ref}}|^{2}
    \\
    \text{subject to} & \qquad \qquad \bm{T}_{\mathrm{min}} \leq \bm{T}_{\mathrm{ref}} \leq \bm{T}_{\mathrm{max}}
\end{split}
\end{align}
Here, $\Lambda$ is the weighting matrix for each joint torque, and $\bm{T}{\mathrm{min}}$ and $\bm{T}_{\mathrm{max}}$ are the minimum and maximum tensions for each wire.
The first term of the objective function minimizes the error between the target joint torque and the actual joint torque, and the second term minimizes the sum of squared wire tensions.

\begin{figure}[t]
  \centering
  \includegraphics[width=1.0\columnwidth]{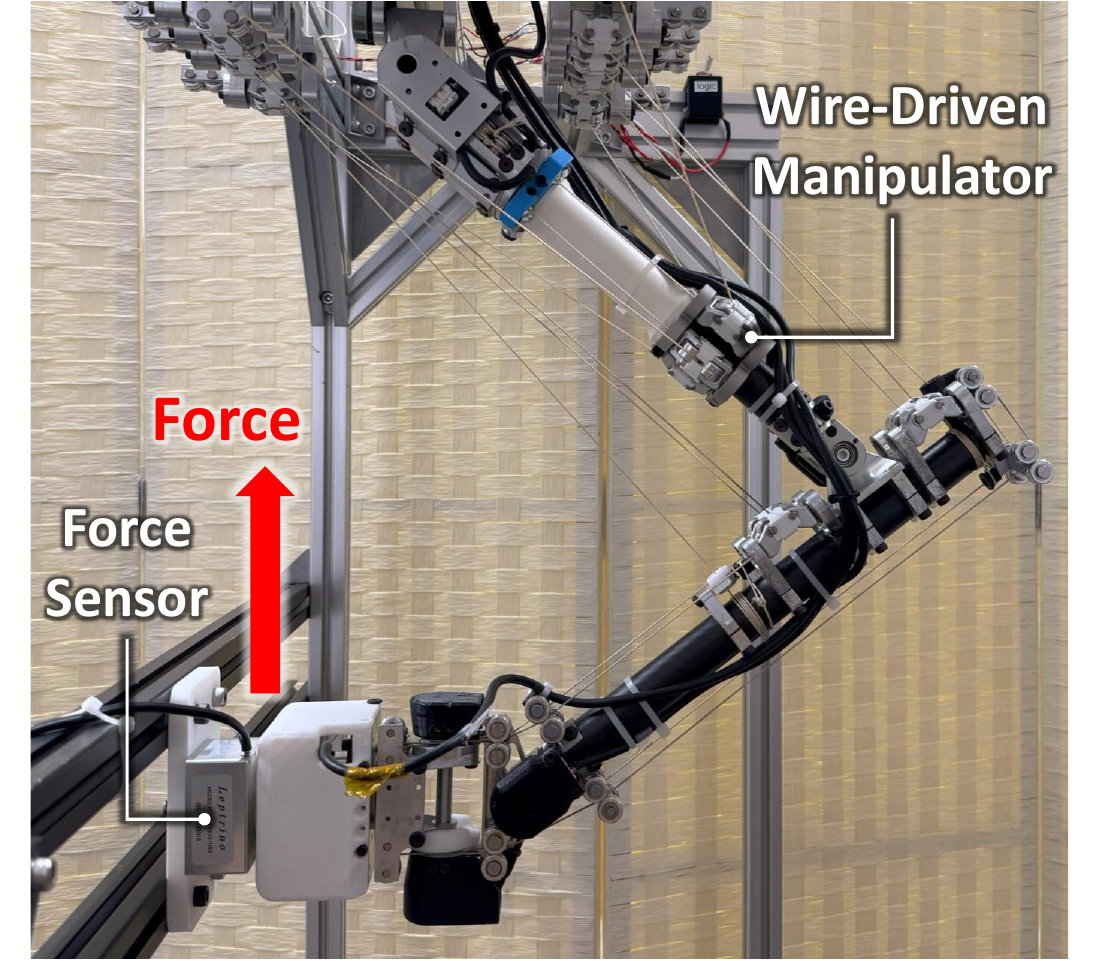}
  \vspace{-2ex}
  \caption{Setup of end-effector force measurement experiment.}
  \label{figure:ee-force-setup}
  \vspace{-1.5ex}
\end{figure}

\begin{figure}[t]
  \centering
  \includegraphics[width=1.0\columnwidth]{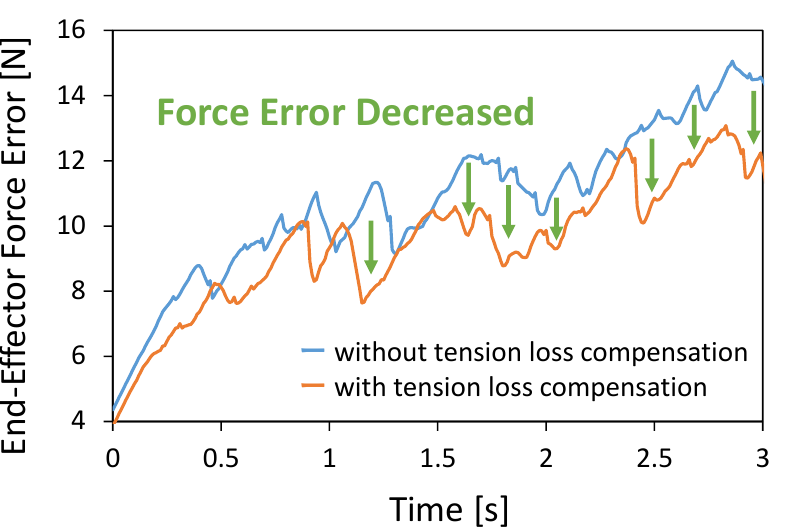}
  \vspace{-2ex}
  \caption{Comparison of end-effector force error with and without tension-loss compensation control.}
  \label{figure:ee-force-plot}
  \vspace{-1.5ex}
\end{figure}

Next, we describe the conversion from the target joint torque $\bm{\tau}{\mathrm{ref}}$ to the target wire tension $\bm{T}{\mathrm{ref}}$ when considering tension loss in the passive pulley.
When tension loss in the passive pulley is taken into account, the following equality holds between the target joint torque $\bm{\tau}{\mathrm{ref}}$ and the target wire tension $\bm{T}{\mathrm{ref}}$:
\begin{align}
  \label{eq:tension-torque-loss}
  \bm{\tau}_{\mathrm{ref}} = -(\bm{\eta} \odot \bm{G})^{\mathrm{T}} \bm{T}_{\mathrm{ref}}
\end{align}
Here, $\bm{\eta}$ is the matrix representing the tension transmission efficiency of each wire at each joint, and is defined as follows.
\begin{align}
  \label{eq:eta}
\bm{\eta} = 
\begin{pmatrix}
  \vspace{2mm}
  \eta_{\mathrm{p}}^{N_{11}} & \eta_{\mathrm{p}}^{N_{12}} & \cdots & \eta_{\mathrm{p}}^{N_{1n}} \\
  \eta_{\mathrm{p}}^{N_{21}} & \eta_{\mathrm{p}}^{N_{22}} & \cdots & \eta_{\mathrm{p}}^{N_{2n}} \\
  \vdots        & \vdots        & \ddots & \vdots        \\
  \eta_{\mathrm{p}}^{N_{m1}} & \eta_{\mathrm{p}}^{N_{m2}} & \cdots & \eta_{\mathrm{p}}^{N_{mn}}
\end{pmatrix},
\quad
\bm{\eta}_{ij} = \eta_{\mathrm{p}}^{N_{ij}},
\quad
\end{align}
Here, $\eta_{\mathrm{p}}$ denotes the tension transmission efficiency per passive pulley, $N_{ij}$ is the number of passive pulleys that wire $i$ has passed through to reach joint $j$, $n$ is the number of joints, and $m$ is the number of wires.
In \eqref{eq:tension-torque-loss}, tension loss during pulley routing is modeled by taking the elementwise (Hadamard) product of the muscle length Jacobian $\bm{G}$ in \eqref{eq:tension-torque} with the tension transmission efficiency matrix $\bm{\eta}$.
By using \eqref{eq:tension-torque-loss}, the target wire tension $\bm{T}_{\mathrm{ref}}$ for compensating tension loss in the passive pulley can be obtained by solving the following quadratic program.
\begin{align}
  \label{eq:calc-tension-eta}
  \begin{split}
    \text{minimize}\quad
      &(\bm{\tau}_{\mathrm{ref}}
      +(\bm{\eta}\odot\bm{G})^{T}\bm{T}_{\mathrm{ref}})^{T}
      \Lambda
      (\bm{\tau}_{\mathrm{ref}}
      +(\bm{\eta}\odot\bm{G})^{T}\bm{T}_{\mathrm{ref}})
    \\
      &\quad +\, |\bm{T}_{\mathrm{ref}}|^{2}
    \\
    \text{subject to}\quad
      &\bm{T}_{\min}\le\bm{T}_{\mathrm{ref}}\le\bm{T}_{\max}
  \end{split}
\end{align}
The target wire tension $\bm{T}_{\mathrm{ref}}$ computed by the above calculations was commanded to the robot to perform end-effector force control.

In this experiment, the vertical upward component of the target end-effector force was increased linearly with time from \SI{0}{\newton} to \SI{40}{\newton}.
The error between the vertical upward component of the target end-effector force and that of the actual end-effector force was measured at each time step.
This trial was performed five times each under conditions with and without tension loss compensation control.

The end-effector force error during the experiment is shown in \figref{figure:ee-force-plot}.
With tension loss compensation control, the root mean square error of the vertical force component was \SI{9.5}{\newton}, whereas it was \SI{10.9}{\newton} without compensation.
These results indicate that tension loss compensation control reduced the end-effector force error by approximately \SI{13}{\percent}.
This experiment confirmed that using data obtained from the wire testing machine can improve the force control accuracy of an actual wire-driven mechanism.
}%
{%
\subsecref{subsec:pulley-efficiency}で計測した受動プーリの張力伝達効率を実際のワイヤ駆動ロボットの力制御に適用する.
実験のセットアップを\figref{figure:ee-force-setup}に示す.
本実験ではワイヤ干渉駆動マニュピレータに目標手先力を指令し, 手先の六軸力センサで目標手先力と実際の手先力の誤差を計測した.
そして受動プーリにおける張力損失を補償した場合としなかった場合の手先力の誤差を比較した.
なお本実験で用いるロボットは直径\SI{1}{\milli\metre}のVectran rope (VB-175, Hayami industry)と, 直径\SI{12}{\milli\metre}の受動プーリを使用している.
本ロボットではワイヤが最大7個の受動プーリを経由する.

本実験で用いた手先力制御則について説明する.
まず目標手先力$\bm{f}_{\mathrm{ref}}$が上位のコントローラーから与えられる.
次に目標手先力$\bm{f}_{\mathrm{ref}}$と, ロボットの手先位置と関節角度の関係を表す関節ヤコビアン$\bm{J}_{\mathrm{r}}$から目標関節トルク$\bm{\tau}_{\mathrm{ref}}$が以下のように求まる.
\begin{align}
  \label{eq:eeforce-torque}
  \bm{\tau}_{\mathrm{ref}} = \bm{J}_{\mathrm{r}}^{\mathrm{T}} \bm{f}_{\mathrm{ref}}
\end{align}

ここからは目標関節トルク$\bm{\tau}_{\mathrm{ref}}$から目標ワイヤ張力$\bm{T}_{\mathrm{ref}}$への変換について説明する.
まず受動プーリにおける張力損失を考慮しない場合の目標ワイヤ張力$\bm{T}_{\mathrm{ref}}$を求める.
目標関節トルク$\bm{\tau}_{\mathrm{ref}}$と目標ワイヤ張力$\bm{T}_{\mathrm{ref}}$の間には以下の等式が成り立つ.
\begin{align}
  \label{eq:tension-torque}
  \bm{\tau}_{\mathrm{ref}} = -\bm{G}^{\mathrm{T}} \bm{T}_{\mathrm{ref}}
\end{align}
ただし, $\bm{G}$は筋長ヤコビアンという行列で, 各関節における各ワイヤのモーメントアームを表す.
特定の関節角度$\bm{q}$における筋長ヤコビアン$\bm{G}(\bm{q})$は以下のように定義される.
\begin{align}
  \label{eq:muscle-jacobian}
  \bm{G}(\bm{q}) = \frac{\partial \bm{l}}{\partial \bm{q}}
\end{align}
ただし$\bm{l}$はワイヤ長さである.
実際のワイヤ駆動ロボットは発揮可能な張力に制限があるため, 以下の二次計画法を解くことで目標ワイヤ張力$\bm{T}_{\mathrm{ref}}$(受動プーリにおける張力損失を考慮しない場合)を求める\cite{kawamura2016jointspace}.
\begin{align}
  \label{calc-tension}
  \begin{split}
    \text{minimize} & \quad (\bm{\tau}_{\mathrm{ref}}+\bm{G}^{T}\bm{T}_{\mathrm{ref}})^{T} \Lambda (\bm{\tau}_{\mathrm{ref}}+\bm{G}^{T}\bm{T}_{\mathrm{ref}})+|\bm{T}_{\mathrm{ref}}|^{2}
    \\
    \text{subject to} & \qquad \qquad \bm{T}_{\mathrm{min}} \leq \bm{T}_{\mathrm{ref}} \leq \bm{T}_{\mathrm{max}}
\end{split}
\end{align}
ただし, $\Lambda$は各関節トルクの重み行列, $\bm{T}_{\mathrm{min}}$ / $\bm{T}_{\mathrm{max}}$ は各ワイヤの最小/最大張力である.
目的関数の第一項は目標関節トルクと実際の関節トルクの間の誤差を最小化する項, 第二項はワイヤ張力の二乗和を最小化する項である.

次に受動プーリにおける張力損失を考慮する場合の目標関節トルク$\bm{\tau}_{\mathrm{ref}}$から目標ワイヤ張力$\bm{T}_{\mathrm{ref}}$への変換について説明する.
受動プーリにおける張力損失を考慮する場合, 目標関節トルク$\bm{\tau}_{\mathrm{ref}}$と目標ワイヤ張力$\bm{T}_{\mathrm{ref}}$の間には以下の等式が成り立つ.
\begin{align}
  \label{eq:tension-torque-loss}
  \bm{\tau}_{\mathrm{ref}} = -(\bm{\eta} \odot \bm{G})^{\mathrm{T}} \bm{T}_{\mathrm{ref}}
\end{align}
ただし$\bm{\eta}$は各関節における各ワイヤの張力伝達効率を表す行列であり, 以下のように定義される.
\begin{align}
  \label{eq:eta}
\bm{\eta} = 
\begin{pmatrix}
  \vspace{2mm}
  \eta_{\mathrm{p}}^{N_{11}} & \eta_{\mathrm{p}}^{N_{12}} & \cdots & \eta_{\mathrm{p}}^{N_{1n}} \\
  \eta_{\mathrm{p}}^{N_{21}} & \eta_{\mathrm{p}}^{N_{22}} & \cdots & \eta_{\mathrm{p}}^{N_{2n}} \\
  \vdots        & \vdots        & \ddots & \vdots        \\
  \eta_{\mathrm{p}}^{N_{m1}} & \eta_{\mathrm{p}}^{N_{m2}} & \cdots & \eta^_{\mathrm{p}}{N_{mn}}
\end{pmatrix},
\quad
\bm{\eta}_{ij} = \eta^{N_{ij}},
\quad
\end{align}
ただし$\eta_{\mathrm{p}}$はプーリ1個あたりの張力伝達効率, $N_{ij}$は関節$j$に至るまでにワイヤ$i$が経由してきた受動プーリの個数, $n$は関節数, $m$はワイヤ本数である.
式\eqref{eq:tension-torque-loss}では, 式\eqref{eq:tension-torque}中の筋長ヤコビアン$\bm{G}$に張力伝達効率$\bm{\eta}$を要素積(アダマール積)することでプーリ経由時の張力損失をモデル化している.
式\eqref{eq:tension-torque-loss}を用いると, 受動プーリにおける張力損失を補償する場合の目標ワイヤ張力$\bm{T}_{\mathrm{ref}}$を以下の二次計画法から求めることができる.
\begin{align}
  \label{eq:calc-tension-eta}
  \begin{split}
    \text{minimize}\quad
      &(\bm{\tau}_{\mathrm{ref}}
      +(\bm{\eta}\odot\bm{G})^{T}\bm{T}_{\mathrm{ref}})^{T}
      \Lambda
      (\bm{\tau}_{\mathrm{ref}}
      +(\bm{\eta}\odot\bm{G})^{T}\bm{T}_{\mathrm{ref}})
    \\
      &\quad +\, |\bm{T}_{\mathrm{ref}}|^{2}
    \\
    \text{subject to}\quad
      &\bm{T}_{\min}\le\bm{T}_{\mathrm{ref}}\le\bm{T}_{\max}
  \end{split}
\end{align}
以上の計算によって求めた目標ワイヤ張力$\bm{T}_{\mathrm{ref}}$をロボットに指令し, 手先力制御を行っている.

本実験では目標手先力の鉛直上向き成分を\SI{0}{\newton}-\SI{40}{\newton}まで時刻に対して線形に増加させた.
そして各タイムステップにおける目標手先力の鉛直上向き成分と実際の手先力の鉛直上向き成分の誤差を計測した.
この試行を張力損失補償制御を導入した条件と導入していない条件のそれぞれで5回ずつ行った。

実験中の手先力の誤差を\figref{figure:ee-force-plot}に示す.
張力損失補償制御をした場合は力の鉛直方向成分の平均二乗誤差の平方根が\SI{9.5}{\newton}だったのに対し, 導入しなかった場合は\SI{10.9}{\newton}であった.
張力損失補償制御をすることで手先力の誤差が約\SI{13}{\percent}小さくなったことが分かる.
この実験からワイヤ試験機で得られたデータを用いることで実際のワイヤ駆動ロボットの力制御の精度を向上できることが確認された.

\begin{figure}[t]
  \centering
  \includegraphics[width=1.0\columnwidth]{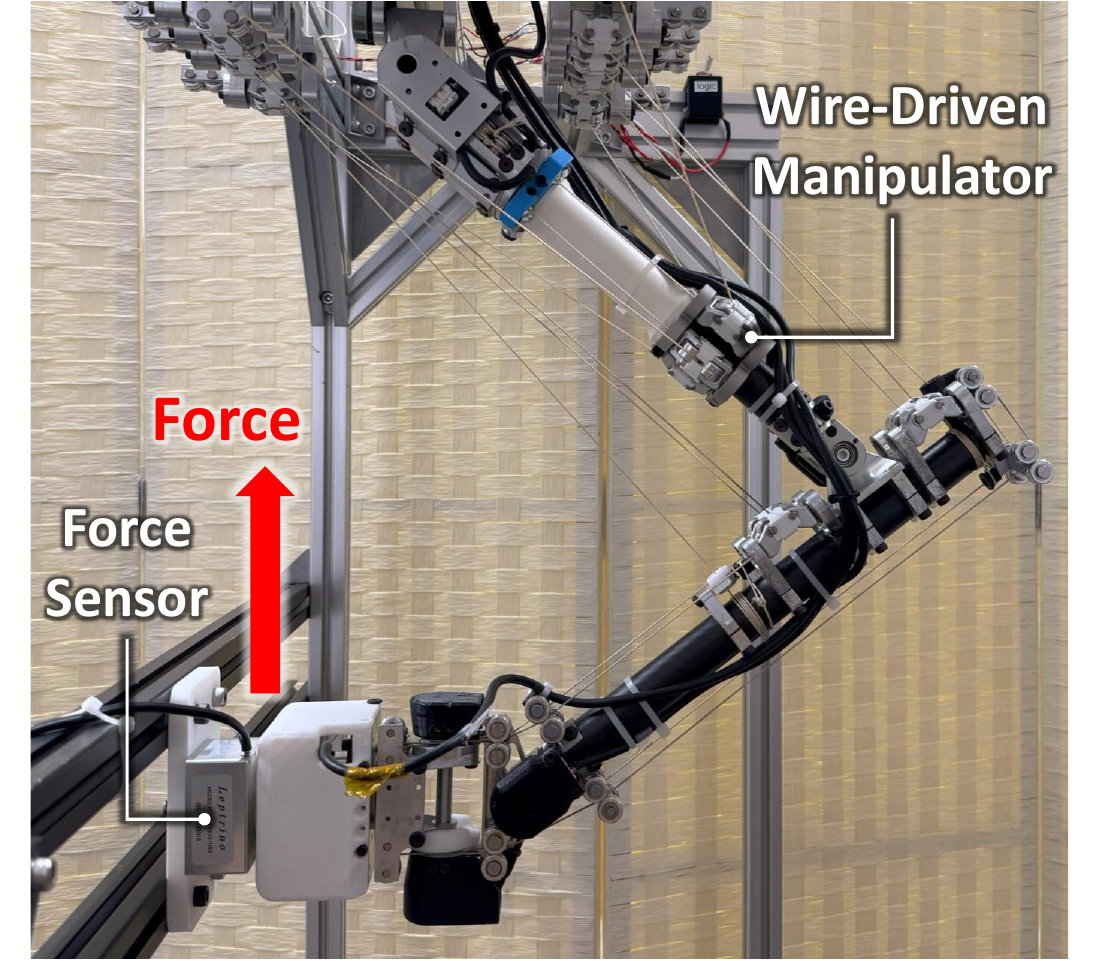}
  \vspace{-2ex}
  \caption{Setup of end-effector force measurement experiment.}
  \label{figure:ee-force-setup}
  \vspace{-1.5ex}
\end{figure}

\begin{figure}[t]
  \centering
  \includegraphics[width=1.0\columnwidth]{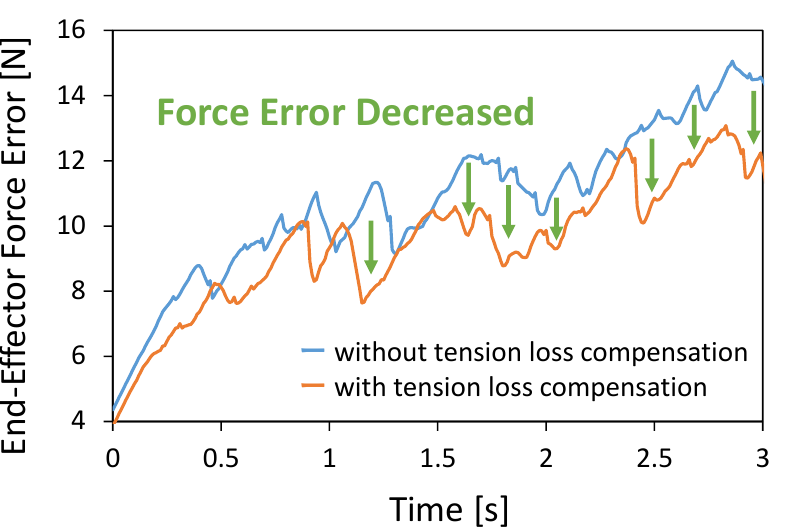}
  \vspace{-2ex}
  \caption{Comparison of end-effector force error with and without tension-loss compensation control.}
  \label{figure:ee-force-plot}
  \vspace{-1.5ex}
\end{figure}
}%

\section{Conclusion} \label{sec:conclusion}
\switchlanguage%
{%
In this paper, we propose a Universal Wire Testing Machine capable of measuring and adjusting wire characteristics for improving the performance of wire-driven robots.
Using this testing machine, we performed initial wire stretch removal, measured the tension transmission efficiency of passive pulley units for eight pulley diameters and four wire types, and measured the dynamic characteristics of variable-length wires.
Furthermore, we used the measured tension transmission efficiency of the passive pulley to improve the accuracy of end-effector force control in an actual wire-driven robot.

Future work includes investigation of the relationship between wire wrap angle around the passive pulley and tension transmission efficiency, and modeling of variable-length wire behavior.
In this study, the wire wrap angle on the passive pulley was fixed at \SI{90}{\degree}, but it is known that variations in wrap angle affect the tension transmission efficiency.
Therefore, by changing not only the pulley diameter and wire type but also the wrap angle to measure tension transmission efficiency, more precise force control of wire-driven robots can be expected.
Additionally, standard elastic and viscous wire models widely used today cannot accurately represent wire behavior when the wire length changes drastically with slack.
Thus, we aim to collect tension data under conditions of large wire length variation using this testing machine and, based on these data, develop a tension-prediction neural network to more accurately predict the behavior of wire-driven robots.
}%
{%
本論文ではワイヤ駆動ロボットの性能向上に向けてワイヤ特性の測定と調節が可能なUniversal Wire Testing Machineを提案した.
本試験機を用いて, ワイヤの初期伸びの除去, 8種類の直径のプーリと4種類のワイヤにおける受動プーリの張力伝達効率の測定, 可変長ワイヤの動特性の測定を行った.
さらに本試験機で得られた受動プーリの張力伝達効率を用いて, 実際のワイヤ駆動ロボットの手先力制御の精度を向上させた.

今後の課題としては, 受動プーリへのワイヤ巻き掛け角度と張力伝達効率の関係の調査と, 可変長ワイヤの挙動のモデル化が挙げられる.
本論文では受動プーリへのワイヤ巻きかけ角度は\SI{90}{\degree}で一定であったが, ワイヤ巻き掛け角度を変化させると張力伝達効率が変化することが知られている.
そのためプーリ直径やワイヤ種類だけでなく巻きかけ角度も変化させて張力伝達効率を測定することで, ワイヤ駆動ロボットの力制御をより高精度にできることが予想できる.
また現在広く使用されている弾性や粘性を用いたワイヤのモデルでは, 緩みを伴ってワイヤの全長が激しく変化する場合のワイヤの挙動を表現することが困難である. 
そこで本試験機を用いてワイヤの全長が激しく変化する場合の張力データを収集し, このデータを元に張力予測ニューラルネットワークを作成することでワイヤ駆動ロボットの挙動をより正確に予測可能となることを目指す.
}%

{
  %\footnotesize
  %\small
  %\bibliographystyle{junsrt}
  \bibliographystyle{IEEEtran}
  \bibliography{main}
}

\end{document}